\documentclass[runningheads]{llncs}

\usepackage{eccv}

\usepackage{eccvabbrv}

\usepackage{graphicx}
\usepackage{booktabs}
\usepackage[accsupp]{axessibility}  

\usepackage{siunitx}
\usepackage{pifont}
\usepackage{array}
\usepackage{subcaption}
\usepackage{multirow}
\usepackage{fontawesome5}

\usepackage{fontawesome5}

\DeclareRobustCommand{\corrauth}{%
\textsuperscript{(\faIcon[regular]{envelope})}%
}

\usepackage{hyperref}

\usepackage{orcidlink}

\begin{document}
\title{FSDC-DETR: A Frequency-Spatial Domain Collaborative DETR for Small Object Detection}
\titlerunning{FSDC-DETR}
\author{
Aiwen Liu\inst{1} \orcidlink{0009-0003-7601-3895} \and
Chengguang Zhu\inst{1}\protect\corrauth \orcidlink{0000-0002-7737-4225} \and
Gang Wang\inst{1} \and
Dandan Zhu\inst{2} \orcidlink{0000-0003-0329-6321} \and
Haodong Lin\inst{1} \orcidlink{0009-0005-3200-190X} \and
Yan Wang\inst{4} \and
Huiyu Zhou\inst{3} \orcidlink{0000-0003-1634-9840} \and
Zhengyi Pan\inst{1}
}

\authorrunning{Liu et al.}

\institute{
Micro-Intelligence, Shanghai 201100, China \\
\email{nevereverinsomnia@gmail.com, chengguang.zhusjtu@gmail.com, \{roy.wang,haodong.lin,eric.pan\}@micro-i.com.cn} \and
East China Normal University, Shanghai 200241, China \\ \email{ddzhu@mail.ecnu.edu.cn} \and
University of Leicester, Leicester LE1 7RH, UK \\ \email{hz143@leicester.ac.uk} \and
Chongqing Normal University, Chongqing 401331, China}

\maketitle

\begingroup
\renewcommand{\thefootnote}{\fnsymbol{footnote}}
\makeatletter
\makeatother
\footnotetext[1]{Aiwen Liu, Chengguang Zhu, and Gang Wang --- Equal contribution.}
\footnotetext[2]{Corresponding author: Chengguang Zhu.}
\endgroup

\begin{abstract}

Small object detection (SOD) remains a challenging task in real-world applications. 
Despite recent advances, existing detectors remain limited by rigid processing that entangle spatial aggregation with implicit frequency aliasing and truncation, leading to inadequate preservation of high-frequency components for SOD. 
To tackle these limitations, we propose a Frequency-Spatial Domain Collaborative Detection Transformer (FSDC-DETR), a novel collaborative framework that explicitly models complementary spatial and frequency representations.
Specifically, we first introduce Dual-Branch Frequency-Spatial Adaptive Fusion (DBFSAF) to enhance frequency diversity and adaptively capture frequency-spatial domain discriminative representations.
Building on these representations, a frequency-spatial interaction scheme is further explored within the hybrid encoder to enable progressive feature propagation to the decoder.
In particular, structure-aware frequency-spatial aggregation is achieved through Shunt Frequency-Spatial Feature Fusion (SFS-FF), establishing bidirectional interaction and progressive cross-scale propagation between frequency and spatial representations for coherent discriminative modeling. 
Meanwhile, informative high-frequency responses are preserved during scale transitions through Frequency-Spatial Dynamic Downsampling (FSD-Down), thereby minimizing frequency degradation throughout multi-scale fusion for the precise SOD.
Experimental results demonstrate that FSDC-DETR achieves state-of-the-art performance, improving AP by 6.4 on VisDrone-DET2019 and 6.6 on AITODv2, with gains of 6.8 and 6.9 AP for small objects. The code is available at \url{https://github.com/nevereverinsomnia/FSDC-DETR}.

\end{abstract}

\section{Introduction}
\label{sec1:introduction}

Object detection is a fundamental task in computer vision that has witnessed significant progress in recent years, supporting a wide range of applications such as industrial defect detection\cite{CHENG2026152_survey_industrial_defect_detection,Yan_2025_CVPR_WPFormer,zhang2023idd,lv2025lightweight}, aerial surveillance\cite{hua2025survey,chen2025yoloms,meng2021conditional,yang2022querydet}, remote sensing\cite{shi2025progressive,11092306_cvpr25,shi2025hs_fpn,li2023large}, medical lesion analysis\cite{fan2024deep_chestx,WEI2026103906,ZHANG2022102415}, and autonomous systems\cite{zhan2024yolopx,sapkota2026object,11297835,11397189}.
Most of these advances have been driven by Convolutional Neural Network (CNN)-based detectors, which typically rely on a series of manually designed components, such as anchor box generation and non-maximum suppression (NMS), and remain heavily dependent on human expertise\cite{fast_rcnn,faster_rcnn,yolov5,yolov8}.
These components often require extensive task-specific tuning, making the overall detection pipeline complex and inefficient in real-world applications. More recently, the end-to-end Transformer-based Detection Transformer (DETR) framework has been introduced to largely alleviate these limitations by reformulating object detection as a direct set prediction problem\cite{DETR_ECCV20}.
Subsequent DETR-based variants have further improved the detection accuracy while significantly improving inference efficiency\cite{RT-DETRv1,RT-DETRv4,DFine_ICLR2025,DEIM}. However, these advances are primarily designed for generic object detection and still face substantial challenges when applied to small object detection (SOD), particularly in scenarios such as industrial defect detection and aerial imagery acquired from unmanned aerial vehicles (UAVs) and satellites.

\begin{figure}[!ht]
\centering
\captionsetup[subfigure]{skip=2pt}

\begin{subfigure}{0.24\linewidth}
    \centering
    \includegraphics[width=\linewidth]{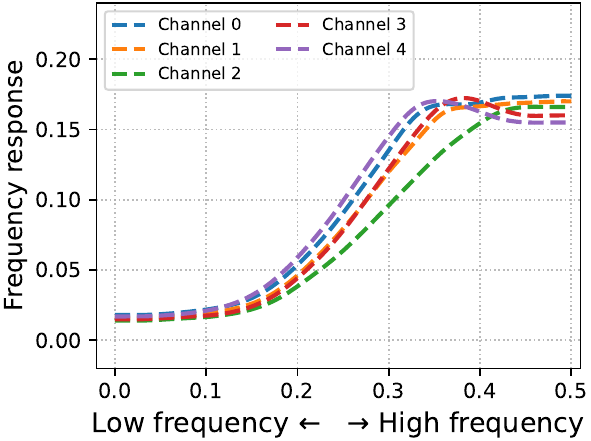}
    \caption{CNN}
    \label{fig1a:cnn_freq}
\end{subfigure}
\hfill
\begin{subfigure}{0.24\linewidth}
    \centering
    \includegraphics[width=\linewidth]{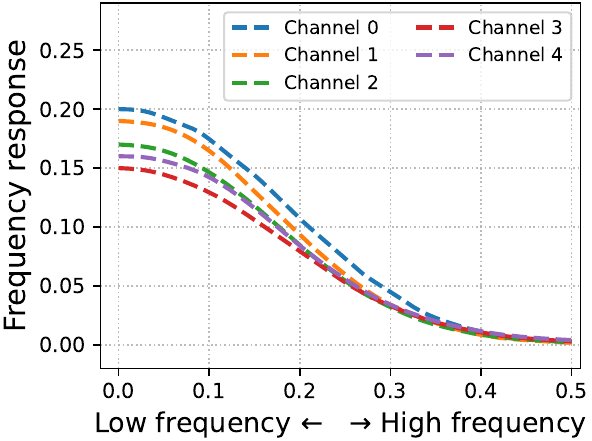}
    \caption{ViT}
    \label{fig1b:vit_freq}
\end{subfigure}
\hfill
\begin{subfigure}{0.24\linewidth}
    \centering
    \includegraphics[width=\linewidth]{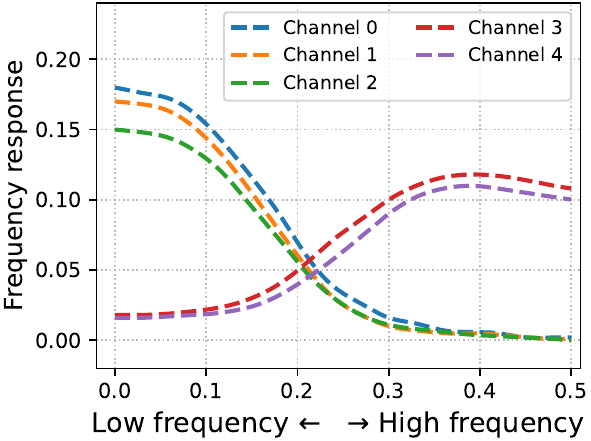}
    \caption{DEIMv2}
    \label{fig1c:deimv2_freq}
\end{subfigure}
\hfill
\begin{subfigure}{0.24\linewidth}
    \centering
    \includegraphics[width=\linewidth]{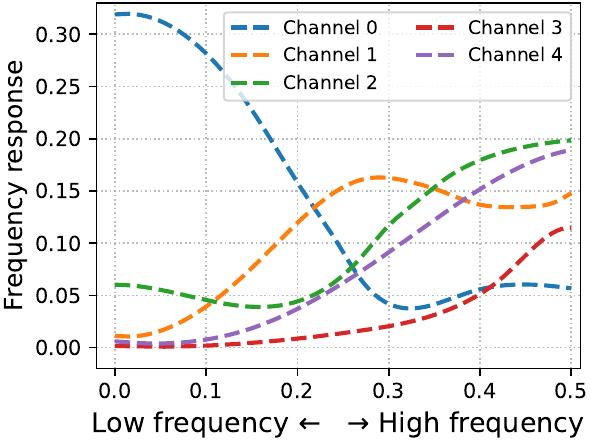}
    \caption{FSDC-DETR}
    \label{fig1d:fsdc_freq}
\end{subfigure}

\caption{Frequency response analysis of different backbone architectures.}
\label{fig1:motivation}
\end{figure}

SOD focuses on identifying and localizing objects with extremely limited spatial extent in images~\cite{AITODv1,AITODv2}. Although spatially limited, these objects play a critical role in real-world applications. 
The challenges of SOD mainly stem from insufficient feature representation capacity in the model backbone, key information loss in the deeper layers of feature hierarchies, and frequency-domain aliasing during multi-scale feature fusion.
As a result, small objects often experience feature loss and inadequate representation, which leads to reduced detection accuracy.
RT-DETR \cite{RT-DETRv1} improves inference efficiency by decoupling intra-scale and cross-scale interactions, yet it does not explicitly model frequency-domain representations. 
As a result, high-frequency details are implicitly attenuated during feature aggregation and downsampling, leading to the loss of fine-grained information critical for SOD tasks.
As illustrated in Fig.~\ref{fig1:motivation}(\subref{fig1a:cnn_freq})-(\subref{fig1b:vit_freq}), CNN and  Vision Transformer (ViT) backbones exhibit substantially different frequency response characteristics~\cite{Chen_2025_ICCV,shu2026waveformer,wang2022antioversmooth,tatsunami2024dynamic-filter,Chenlinwei_FDConv_CVPR25,paul2022vision,9522696,Wang_CNN_high_freq_cvpr20}.
Although DEIMv2~\cite{DEIMv2} adopts a dual-branch CNN-ViT backbone, feature fusion is performed via direct concatenation followed by a simple channel projection.
In the absence of explicit frequency-aware modeling, this design overlooks the heterogeneous spectral properties of CNN and ViT representations, leading to insufficient preservation of fine-grained high-frequency details, as illustrated in Fig.~\ref{fig1:motivation}(\subref{fig1c:deimv2_freq}).

In this paper, we introduce the Frequency-Spatial Domain Collaborative Detection Transformer (FSDC-DETR), a novel end-to-end transformer-based object detection framework designed to enhance SOD through explicit frequency-spatial modeling. 
To construct frequency-spatial collaborative representations and mitigate frequency aliasing and truncation in dual-branch backbones, we first propose a Dual-Branch Frequency-Spatial Adaptive Fusion (DBFSAF) module that enhances frequency diversity of the dual-branch representations, adaptively fuses and selects features with increased preservation of high-frequency components, and performs Partial Frequency-Spatial Refinement (PFSR). The corresponding frequency response is illustrated in Fig.~\ref{fig1:motivation}(\subref{fig1d:fsdc_freq}).
Building upon these refined representations, we further develop a Shunt Frequency-Spatial Feature Fusion (SFS-FF) strategy within a hybrid encoder to propagate frequency-aware features across scales. By decoupling and collaboratively refining spatial and frequency components, SFS-FF enables structure-aware cross-scale aggregation without suppressing fine-grained details.
To preserve frequency integrity during resolution transitions in multi-scale modeling, we introduce a Frequency-Spatial Dynamic Downsampling (FSD-Down) operator based on learnable wavelet decomposition and grouped convolution. FSD-Down retains informative high-frequency responses while maintaining spatial consistency, ensuring minimal frequency degradation throughout the feature hierarchy.

Experimental results on VisDrone-DET2019 \cite{VisDrone-DET2019} and AITODv2 \cite{AITODv2} demonstrate that FSDC-DETR establishes a new state-of-the-art for SOD, delivering substantial improvements over strong baselines, especially in the AP$_S$ metric. These results verify that explicit frequency-spatial modeling effectively preserves high-frequency cues critical for SOD.
In summary, our main contributions are as follows:
\begin{itemize}
\item We propose FSDC-DETR, a frequency-spatial domain collaborative DETR framework that explicitly models complementary spatial and frequency representations to address aliasing and high-frequency degradation in SOD.

\item To systematically construct and propagate frequency-aware representations, we design a unified frequency-spatial collaborative pipeline composed of three progressively integrated components. 
(i) The DBFSAF module mitigates frequency aliasing and truncation phenomenon in dual-branch backbones by enhancing spectral diversity and adaptively preserving informative high-frequency cues. 
(ii) The SFS-FF strategy within the hybrid encoder propagates frequency-enriched features across scales through structure-aware frequency-spatial aggregation. 
(iii) The FSD-Down operator, built upon learnable wavelet decomposition and grouped convolution, preserves frequency integrity during scale transitions while maintaining spatial consistency.

\item Experimental results on VisDrone-DET2019 and AITODv2 demonstrate state-of-the-art performance, improving AP by 6.4 and 6.6, with gains of 6.8 and 6.9 AP on small objects, respectively.
\end{itemize}

\section{Related Work}
\label{sec2:Related Work}
\subsection{Small Object Detection}
\label{subsec2.1:Small Object Detection}

SOD remains particularly challenging due to limited pixel support, low signal-to-noise ratio, and severe feature attenuation during hierarchical structure. 
Compared with the general detection scenario, small objects occupy only a few spatial locations, making them highly vulnerable to the spatial reduction and the smoothing representation. 
Repeated multi-scale aggregation introduces progressive attenuation of high-frequency components, undermining the discriminative signals required for accurate detection.

Conventional CNN-based detectors, such as Fast R-CNN \cite{fast_rcnn,faster_rcnn} and YOLO\cite{yolov5,yolov8,yolov9,yolov10,yolo11_ultralytics,yolov12,yolov13,yolo26_ultralytics,FBRT_YOLO_AAAI25,mamba_yolo,wang2025yoloe}, often exhibit performance degradation on small objects due to insufficient fine-grained feature representation and limited long-range dependency modeling. Although data augmentation strategies (e.g., copy-paste techniques\cite{copy_paste}) and IoU-aware loss \cite{zhang2022focaleiou,yang2025pinwheel,alpha_iou,tong2023wiseiou} formulations attempt to mitigate scale imbalance, these approaches primarily operate at the data or optimization level, leaving representation degradation comparatively underexplored\cite{zhang2022dino,RT-DETRv4,DEIM}.

Transformer-based detectors, particularly DETR \cite{DETR_ECCV20} and its variants\cite{RT-DETRv1,RT-DETRv2,RT-DETRv3,RT-DETRv4,DEIM,DEIMv2,RF_DETR,zhang2025mrdetr,DFine_ICLR2025,REN2026103728_FII-DETR,li2023lite_detr,zhao2024msdetr}, introduce global attention mechanisms that enhance long-range modeling and remove hand-crafted components such as anchors and NMS. However, vanilla DETR suffers from high computational cost and slow convergence, limiting its applicability in real-time scenarios. Recent variants, including RT-DETR\cite{RT-DETRv1,RT-DETRv2,RT-DETRv3,RT-DETRv4}, D-FINE \cite{DFine_ICLR2025}, Mr. DETR \cite{zhang2025mrdetr}, and DEIM\cite{DEIM}, improve efficiency and accelerate training, making DETR-style models more practical. 
Nevertheless, despite architectural refinements, existing hybrid encoder designs still rely heavily on conventional multi-scale fusion and spatial downsampling schemes, where high-frequency information may be weakened during scale transition, posing challenges for SOD\cite{zhang2025uav_detr,shi2025hs_fpn,set_2025_CVPR}.

\subsection{DETRs with Vision Foundation Model}
\label{subsec2.2:End-to-end object detectors}

Vision foundation models, such as the DINO family \cite{caron2021dinov1,oquab2023dinov2,simeoni2025dinov3}, SAM3 \cite{carion2025sam3segmentconcepts}, and C-RADIOv4 \cite{ranzinger2026cradiov4techreport}, have substantially alleviated the limitations of feature representation in diverse visual scenarios.
RT-DETRv4 \cite{RT-DETRv4} introduces a semantic transfer strategy that leverages DINOv3 to supervise the AIFI\cite{RT-DETRv1} module. Specifically, the high-level representations produced by DINOv3 are used as supervisory signals for the AIFI output to enhance semantic richness and improve feature discrimination.
RF-DETR \cite{RF_DETR} replaces conventional backbones with a DINOv2-pretrained ViT to inherit strong self-supervised visual priors and further employs weight-sharing neural architecture search to optimize detection performance under different computational budgets.
DEIMv2 \cite{DEIMv2}, built upon DEIM \cite{DEIM}, adopts a dual-branch architecture that couples DINOv3 with a lightweight CNN backbone. Although this design aims to integrate global contextual modeling with local structural extraction, how to effectively facilitate the interaction between the two branches still deserves further exploration.
In particular, DEIMv2 overlooks the frequency-domain disparity between transformer-based and CNN-based representations. ViTs typically exhibit low-pass filtering characteristics due to global attention aggregation\cite{Chen_2025_ICCV,shu2026waveformer,wang2022antioversmooth,tatsunami2024dynamic-filter}, whereas CNNs inherently preserve stronger high-frequency components\cite{Chenlinwei_FDConv_CVPR25,paul2022vision,9522696,Wang_CNN_high_freq_cvpr20}. Without explicit frequency-aware alignment, directly fusing these heterogeneous representations may introduce aliasing effects and hinder effective information complementarity. Consequently, the potential synergy between the two branches is not fully exploited.

\subsection{Frequency Domain Learning}
\label{subsec2.3:Frequency domain learning}

Frequency-domain analysis has long served as a fundamental tool in signal processing \cite{gonzalez2009digital,pitas2000digital}.
Recently, it has been increasingly incorporated into deep learning, where it has influenced model robustness \cite{yin2019fourier} and the generalization capability of computer vision models \cite{Chenlinwei_FDConv_CVPR25,shi2025hs_fpn,Wang_CNN_high_freq_cvpr20}.
Beyond theoretical analysis, frequency-aware designs have been integrated into network architectures to enhance global representation learning\cite{sun2024FESL,Li_SFSConv_CVPR25,Li_2025_ICCV_waveseg,Yan_2025_CVPR_WPFormer,TIP25_Spatial_Frequency_mamba,huang2023adaptive} and domain generalization\cite{li2025noise_CVPR25,Wang_CNN_high_freq_cvpr20}.
Several works explicitly incorporate frequency modeling into CNNs. For instance, FcaNet\cite{qin2021fcanet} models channel attention in the frequency domain, extending global average pooling to multi-spectral channel representations.
HS-FPN\cite{shi2025hs_fpn} employs high-pass filtering to capture high-frequency responses and uses them as spatial- and channel-wise masks within FPN to enhance fine-grained representations for SOD.
FreqFusion\cite{FreqFusion} introduces a frequency-aware feature fusion framework that employs adaptive low-pass and high-pass filtering to reduce intra-class inconsistency and enhance boundary details during multi-scale feature fusion. 
SFM\cite{SFM_BIT_Chenlinwei} reduces aliasing by modulating high-frequency features before downsampling and restoring them during upsampling to better preserve fine details. FADC\cite{FADC_BIT_Chenlinwei} improves dilated convolution from a spectral perspective by adaptively adjusting dilation rates and reweighting frequency components to better balance receptive field size and effective bandwidth.
SET~\cite{set_2025_CVPR} advances tiny object detection from a spectral perspective by suppressing background-dominated high-frequency noise and introducing adversarial perturbation learning to enhance the saliency of tiny-object representations.
Recent studies \cite{park2022how,Zhou_2017_CVPR} demonstrate that capturing balanced representations of both high- and low-frequency components can enhance model performance. FCENet\cite{FCENet_CVPR25} exploits cross-field frequency correlations between high- and low-frequency components, dynamically selecting and fusing complementary spectral information for improved NIR-assisted image denoising. Nevertheless, existing frequency-aware methods mostly focus on frequency modeling within individual architectures, leaving the complementary fusion of high-frequency details~\cite{Chen_2025_ICCV,shu2026waveformer,wang2022antioversmooth,tatsunami2024dynamic-filter} and low-frequency semantics~\cite{Chenlinwei_FDConv_CVPR25,paul2022vision,9522696,Wang_CNN_high_freq_cvpr20} remains insufficiently investigated.

\section{The Proposed Model} 
\label{sec2: The Proposed Model}
\subsection{The FSDC-DETR Overview}
\label{subsec3.1:overview}

\begin{figure}[!ht]
    \centering
    \includegraphics[width=1\linewidth]{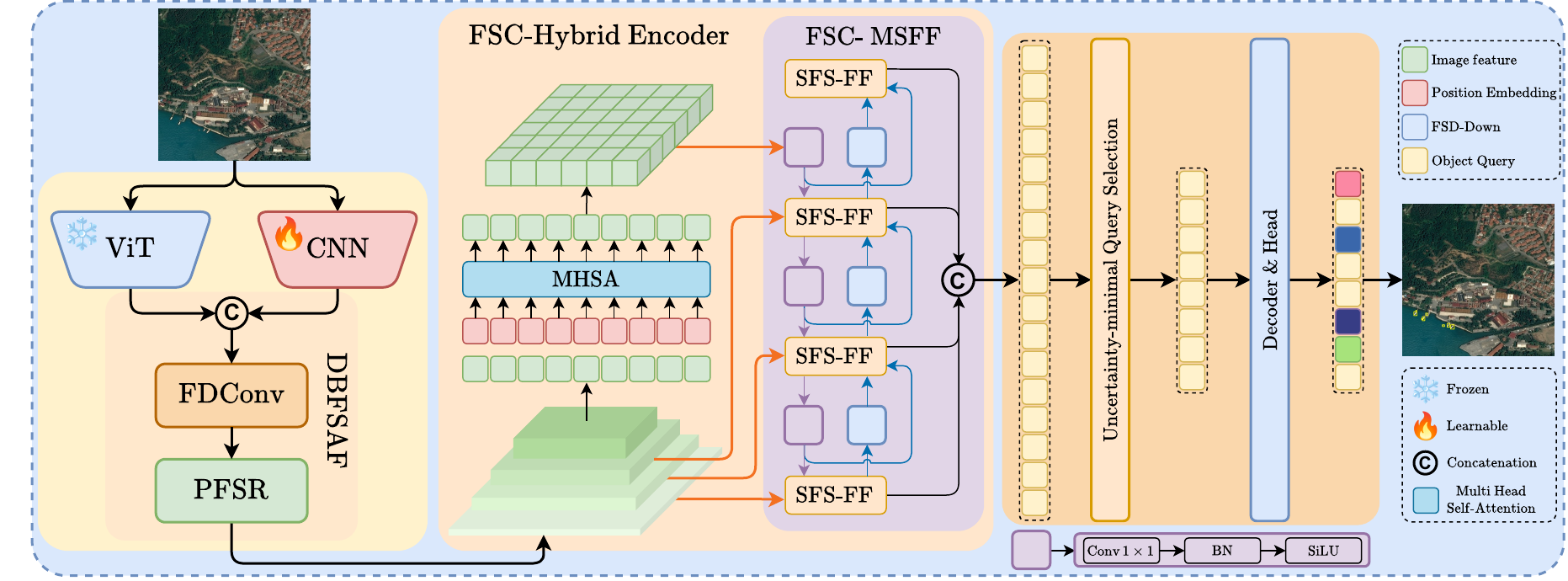}
    \caption{Overall architecture of the proposed FSDC-DETR}
    \label{fig2:overviw}
\end{figure}

As illustrated in Fig.~\ref{fig2:overviw}, we propose FSDC-DETR, an enhanced variant of DEIMv2 \cite{DEIMv2} for SOD through explicit frequency-spatial modeling. The overall framework forms a unified frequency-spatial collaborative pipeline that progressively constructs, propagates, and preserves frequency-aware representations.
Specifically, to construct frequency-aware representations, we introduce the DBFSAF module which enhances spectral diversity, adaptively preserves complementary high-frequency cues, and performs frequency-spatial collaborative refinement. 
Building upon these refined representations, the SFS-FF module propagates frequency-enriched features across scales via structure-aware spectral-spatial aggregation. To preserve frequency integrity during resolution transitions, the FSD-Down module facilitates scale transformation while retaining informative high-frequency responses and maintaining spatial consistency.

\subsection{Dual-Branch Frequency-Spatial Adaptive Fusion}
\label{subsec3.2:DBFSAF}

\begin{figure}[ht]
    \centering
    \includegraphics[width=0.8\linewidth]{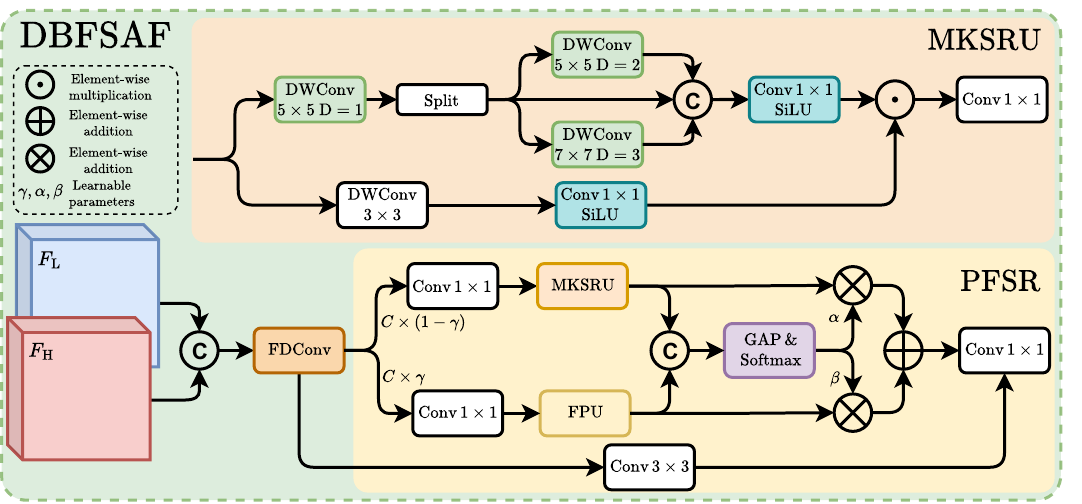}
    \caption{Illustration of the Dual-Branch Frequency-Spatial Adaptive Fusion.}
    \label{fig3: DBFSAF}
\end{figure}

The representational capabilities of CNNs and ViTs have been widely demonstrated.
ViTs exhibit a pronounced low-pass filtering behavior \cite{Chen_2025_ICCV,shu2026waveformer,wang2022antioversmooth,tatsunami2024dynamic-filter}, primarily caused by the global attention mechanism, which tends to over-smooth feature representations and weakens local modeling capability. In contrast, CNNs naturally preserve high-frequency components \cite{Chenlinwei_FDConv_CVPR25,paul2022vision,9522696,Wang_CNN_high_freq_cvpr20} due to their local receptive fields, but often suffer from limited global context modeling. 
DEIMv2 \cite{DEIMv2} introduces a dual-branch backbone composed of a DINOv3-pretrained ViT \cite{simeoni2025dinov3}, which provides robust feature representations in various scenarios, and a simple CNN, enabling complementary representation learning and achieving state-of-the-art performance in general object detection.
However, DEIMv2 combines heterogeneous representations from the ViT and CNN branches via channel-wise concatenation, followed by a 1×1 convolution prior to multi-scale aggregation. Without explicit frequency-aware adaptation, such a design does not account for the distinct frequency characteristics of ViTs and CNNs, potentially limiting the effective preservation and utilization of high-frequency components, particularly for SOD.

The detection task is crucial to adaptively capture diverse frequency components across the feature map, as different spatial locations exhibit distinct local frequency characteristics. In particular, object boundaries and fine structures rely heavily on high-frequency information for precise localization, whereas homogeneous or smooth regions benefit less from such details.

To address the aforementioned limitation, we introduce a DBFSAF module, inspired by Frequency Dynamic Convolution (FDConv) \cite{Chenlinwei_FDConv_CVPR25} and Space-Frequency Selection Convolution (SFS-Conv) \cite{Li_SFSConv_CVPR25}. As depicted in Fig.~\ref{fig3: DBFSAF},
DBFSAF first applies FDConv to the concatenated dual-branch features extracted from HGNetv2\cite{cui2021hgnetv2} and DINOv3\cite{simeoni2025dinov3}, enabling dynamic frequency diversification while maintaining discriminative high-frequency components within the adaptively selected feature representations. 
Subsequently, a PFSR module performs collaborative frequency-spatial refinement to enhance representation quality with minimal computational overhead.
Let $F^{P}_L$ and $F^{P}_H$ denote the low-frequency and high-frequency feature maps from DINOv3 and HGNETv2, at stage $\mathbf{P} \in \{2,3,4,5\}$, respectively. The DBFSAF module can be formulated as:
\begin{equation}
    \label{eq:1}
    F^{P}_{\mathbf{C}},F^{P}_{\mathbf{FS}} = \operatorname{Partial_{\gamma}}\left(\operatorname{FDConv}\left( \operatorname{Concat}\left[\mathbf{F}^{P}_{\mathrm{L}}, \mathbf{F}^{P}_{\mathrm{H}}\right]\right)\right),
\end{equation}
where $\operatorname{Partial_{\gamma}}\left(\cdot\right)$ denotes a channel partition operation applied to the FDConv output with $C$ channels. Specifically, the feature map is divided according to a hyperparameter $\gamma$, which yields $C\times \gamma$ channels feature $F^{P}_{FS}$ for jointly Frequency-Spatial refinement and $C\times \left(1-\gamma\right)$ channels feature $F^{P}_{C}$ preserved to alleviate channel redundancy. This partial design reduces redundant inter-channel interactions and avoids unnecessary computations, thereby achieving a favorable trade-off between robustness and accuracy \cite{huang2025partialchannelnetworkcompute_AAAI26}. 

Building upon the partially selected representation $F_{\mathbf{FS}}$, we further introduce a Multi-Kernel Spatial Refine Unit (MKSRU) that operates in coordination with the Frequency Processing Unit (FPU) proposed in \cite{Li_SFSConv_CVPR25}, to jointly enhance spatial structures and frequency-sensitive responses. The refinement process is formulated as follows:

\begin{gather}
F^{P}_{\mathbf{S}}
=
\operatorname{MKSRU}\!\left(
\mathrm{Conv}_{1\times1}
\left(F^{P}_{\mathbf{FS}}\right)
\right),
\label{eq:2}
\\[6pt]
F^{P}_{\mathbf{F}}
=
\operatorname{FPU}\!\left(
\mathrm{Conv}_{1\times1}
\left(F^{P}_{\mathbf{FS}}\right)
\right),
\label{eq:3}
\\[6pt]
\hat{F}^{P}_{\mathbf{S}}
=
\alpha \odot
\operatorname{Softmax}\!\left(
\operatorname{GAP}\!\left(
\operatorname{Concat}\!\left[
F^{P}_{\mathbf{S}},\;
F^{P}_{\mathbf{F}}
\right]
\right)
\right)
\odot
F^{P}_{\mathbf{S}},
\label{eq:4}
\\[6pt]
\hat{F}^{P}_{\mathbf{F}}
=
\beta \odot
\operatorname{Softmax}\!\left(
\operatorname{GAP}\!\left(
\operatorname{Concat}\!\left[
F^{P}_{\mathbf{S}},\;
F^{P}_{\mathbf{F}}
\right]
\right)
\right)
\odot
F^{P}_{\mathbf{F}},
\label{eq:5}
\\[6pt]
F^{P}
=
\mathrm{Conv}_{1\times1}\!\left(
\operatorname{Concat}\!\left[
\hat{F}^{P}_{\mathbf{S}}
+
\hat{F}^{P}_{\mathbf{F}},\;
\mathrm{Conv}_{3\times3}
\left(F^{P}_{\mathbf{C}}\right)
\right]
\right),
\label{eq:6}
\end{gather}
where $\odot$ denotes the Hadamard product, $\alpha$ and $\beta$ are learnable scalar parameters, and $\operatorname{GAP}\left(\cdot\right)$ denotes global average pooling. $\hat{F}^{P}_\mathbf{S}$ and $\hat{F}^{P}_\mathbf{F}$ represent the channel-attended spatial and frequency refinement features $F^{P}_\mathbf{S}$ and $F^{P}_\mathbf{F}$ generated by MKSRU and FPU, respectively. 

Multi-kernel convolution blocks have been widely adopted in various computer vision tasks to enhance and refine spatial representations\cite{rahman2025mkunet,sun2024FESL,zhang2025decoding_aaai26,liu2025lidar_ACM_MM25}. Motivated by their strong capability in capturing multi-scale structural cues, we design the developed MKSRU as follows. Let the input feature be $X_{\mathbf{in}}\in\mathbb{R}^{ \mathbf{C}\times\mathbf{H}\times\mathbf{W}}$:

\begin{gather}
X^{\frac{1}{2}\mathbf{C}}_{\mathbf{D}},\;
X^{\frac{1}{8}\mathbf{C}}_{\mathbf{D}},\;
X^{\frac{3}{8}\mathbf{C}}_{\mathbf{D}}
=
\operatorname{Split}\!\left(
\operatorname{DWConv}^{(1)}_{5\times5}
\left(X_{\mathbf{in}}\right)
\right),
\label{eq:7}
\\[6pt]
\hat{X}_{\mathbf{D}}
=
\operatorname{Concat}\!\left[
\operatorname{DWConv}^{(3)}_{7\times7}
\left(X^{\frac{1}{2}\mathbf{C}}_{\mathbf{D}}\right),\;
\operatorname{DWConv}^{(2)}_{7\times7}
\left(X^{\frac{3}{8}\mathbf{C}}_{\mathbf{D}}\right),\;
X^{\frac{1}{8}\mathbf{C}}_{\mathbf{D}}
\right],
\label{eq:8}
\\[6pt]
X_{\mathbf{out}}
=
\operatorname{SiLU}\!\left(
\mathrm{Conv}_{1\times1}
\left(\hat{X}_{\mathbf{D}}\right)
\right)
\odot
\operatorname{SiLU}\!\left(
\mathrm{Conv}_{1\times1}
\left(
\mathrm{Conv}_{3\times3}
\left(X_{\mathbf{in}}\right)
\right)
\right),
\label{eq:9}
\end{gather}
where $\operatorname{DWConv}^{\left(i\right)}_{j \times j}$ denotes a depth-wise convolution with kernel size $j\times j$ and dilation rate $i$, and $\operatorname{SiLU}\left(\cdot\right)$ denotes the activation function.

The proposed DBFSAF block explicitly leverages the complementary frequency characteristics of ViT and CNN representations. Through FDConv and PFSR module, the framework increases frequency diversity while preserving discriminative high-frequency details and suppressing redundant channel interactions. The coordinated frequency-spatial enhancement via MKSRU and FPU further enables multi-scale structural aggregation with minimal computational overhead. The DBFSAF is particularly advantageous for SOD, where fine-grained boundary cues are essential, leading to improved localization accuracy and robustness.

\subsection{Shunt Frequency-Spatial Feature Fusion}
\label{subsec3.3:SFS-FF}

\begin{figure}
    \centering
    \includegraphics[width=0.8\linewidth]{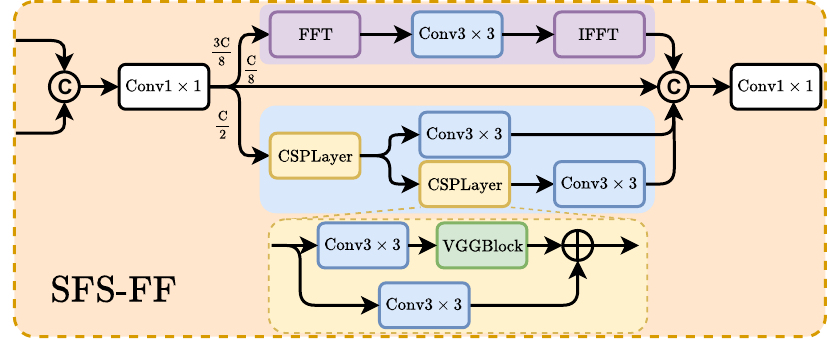}
    \caption{Illustration of the Shunt Frequency-Spatial Feature Fusion.}
    \label{fig4:Fusion}
\end{figure}

Most existing hybrid encoders \cite{DEIM,DEIMv2,RT-DETRv1,RT-DETRv2,RT-DETRv3,RT-DETRv4,DFine_ICLR2025} rely on convolution-based feature fusion and downsampling operations with spatial reduction, which often incur frequency misalignment and attenuate high-frequency components that are essential for SOD. 
Motivated by this limitation, we revisit the hybrid encoder and propose the Frequency-Spatial Collaborative Hybrid Encoder (FSC-Hybrid Encoder) module jointly the SFS-FF and FSD-Down, which is introduced in Section~\ref{subsec3.4:FSD_Down}. This design realizes Frequency-Spatial Collaborative Multi-Scale Feature Fusion (FSC-MSFF) by explicitly modeling complementary frequency and spatial representations. By collaboratively refining frequency cues and enhancing spatial structure awareness, the proposed hybrid encoder yields more discriminative fine-grained features for SOD.

As illustrated in Fig.~\ref{fig4:Fusion}, given the input features
$F_{\mathbf{in}} \in \mathbb{R}^{C \times H \times W}$ and
$\hat{F}_{\mathbf{in}} \in \mathbb{R}^{C \times 2H \times 2W}$, the SFS-FF module is formulated as:

\begin{gather}
F^{}_{S},\, F_{F},\, \acute{F}
= \mathrm{Conv}_{1\times 1} \!\left(
\operatorname{Concat}\!\left[
F_{\mathbf{in}},\;
\operatorname{Down}\!\left(\hat{F}_{\mathbf{in}}\right)
\right]
\right),
\label{eq:10}
\\[6pt]
F_{\mathbf{out}}
= \mathrm{Conv}_{1\times 1} \!\left(
\operatorname{Concat}\!\left[
\operatorname{IFFT}\!\left(
\mathrm{Conv}_{3\times 3}\!\left(
\operatorname{FFT}(F_{S})
\right)
\right),
\operatorname{SRM}(F_{F}),
\acute{F}
\right]
\right),
\label{eq:11}
\end{gather}
where $F_{S}$, $F_{F}$, and $\acute{F}$ denote the outputs of the $1\times1$ convolution, with channel allocations of $\frac{C}{2}$ for spatial refinement, $\frac{3C}{8}$ for frequency refinement, and $\frac{C}{8}$ for residual connection, respectively.
$\operatorname{FFT}\left(\cdot\right)$ and $\operatorname{IFFT}\left(\cdot\right)$ denote the fast Fourier transform and the inverse fast Fourier transform, respectively.
$\operatorname{SRM}\left(\cdot\right)$ denotes spatial refine module \cite{DEIMv2, RT-DETRv1},
$\operatorname{Down}(\cdot)$ represents the proposed FSD-Down.

\subsection{Frequency-Spatial Dynamic Downsampling}
\label{subsec3.4:FSD_Down}

\begin{figure}
    \centering
    \includegraphics[width=0.9\linewidth]{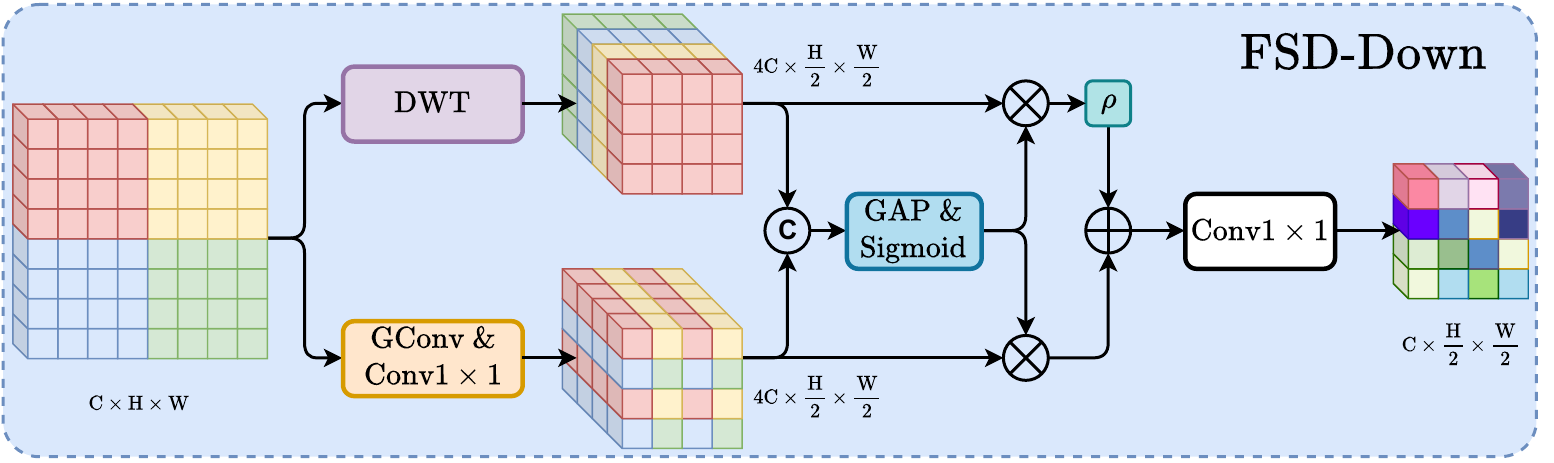}
    \caption{Illustration of the Frequency-Spatial Dynamic Downsampling.}
    \label{fig5:FSD-Down}
\end{figure}

Conventional multi-scale feature fusion \cite{RT-DETRv1,RT-DETRv2,RT-DETRv3,RT-DETRv4,DEIM,DEIMv2} predominantly relies on convolution-based downsampling for spatial reduction, which inherently couples spatial aggregation with implicit frequency aliasing. 
Such operations introduce aliasing artifacts and attenuate high-frequency responses that are essential for the precise SOD. 
UAV-DETR \cite{zhang2025uav_detr} proposed a frequency focused downsampling module that applies average pooling followed by a fast Fourier transform to reduce spatial resolution. However, average pooling fundamentally behaves as a low-pass filter, smoothing local variations and suppressing high-frequency components prior to frequency transformation. As a result, fine-grained discriminative details may already be degraded before frequency modeling is applied.
Therefore, an effective downsampling strategy should explicitly preserve informative high-frequency cues while maintaining spatial structural consistency during scale transition.

Building upon these observations and prior HWD \cite{hwd_xu_pr23} designs, we develop the FSD-Down module, as illustrated in Fig.~\ref{fig5:FSD-Down}, which leverages grouped convolution to parameterize wavelet-based decomposition.
Specifically, the module employs grouped convolution to dynamically calibrate sub-band responses while preserving spatial structural alignment across feature levels.
Given an input feature map $F_{\mathbf{in}} \in \mathbb{R}^{C \times H \times W}$, FSD-Down can be formulated as:
\begin{gather}
F_{\textbf{CA}} 
= \sigma \!\left(
\rm {GAP} \!\left(
\rm {Concat} \!\left[
\rm {DWT}(F_{\mathbf{in}}),\;
\rm {Conv_{1\times 1}}\left(\rm {GConv}(F_{\mathbf{in}})
\right)\right]
\right)
\right),
\label{eq:12} \\[6pt]
F_{\textbf{out}} 
= \rm {Conv}_{1\times 1} \!\left(
\rho \odot F_{\textbf{CA}} \odot 
\rm {DWT}(F_{\mathbf{in}})
+ F_{\textbf{CA}} \odot 
\rm {Conv_{1\times 1}}\left(\rm {GConv}(F_{\mathbf{in}})
\right)\right),
\label{eq:13}
\end{gather}
where $\operatorname{DWT}(\cdot)$ denotes the 2D discrete wavelet transform that decomposes the input into four sub-bands $\{F_{\mathbf{LL}}, F_{\mathbf{LH}}, F_{\mathbf{HL}}, F_{\mathbf{HH}}\}$ with spatial resolution $\frac{H}{2}\times\frac{W}{2}$, corresponding to the low-frequency approximation and horizontal, vertical, and diagonal high-frequency details, respectively.
$\operatorname{GConv}\left(\cdot \right)$ represents a group convolution with kernel size and stride of 2 and $C$ groups , $\operatorname{GAP}$ is the global average pooling, and $\sigma\left(\cdot\right)$ denotes the Sigmoid activation function. The parameter $\rho$ is a learnable scaling factor that adaptively modulates frequency responses, enabling the model to emphasize frequency patterns and local contrasts that are most discriminative for SOD.

\section{Experiments}
\label{sec4:experiments}

\subsection{Datasets}
\label{subset4.1: Datasets}

We perform quantitative evaluations on two widely adopted SOD benchmarks, VisDrone-DET2019 \cite{VisDrone-DET2019} and AITODv2 \cite{AITODv2}.

VisDrone-DET2019 consists of 10,209 images captured by UAVs, including 6,471 images for training, 548 for validation, and 3,190 for testing. The dataset provides bounding-box (BBox) annotations for 10 object categories and spans a wide range of scene densities, from sparse environments to highly congested scenarios. The distribution of object instances exhibits a heavy-tailed pattern, with an average of 40.7 objects per image and a standard deviation of 46.4, posing significant challenges for accurate detection.

AITODv2 contains 28,036 aerial images with a total of 752,745 annotated instances, divided into 11,214 training images, 2,804 validation images, and 14,018 testing images. The dataset provides BBox annotations for 8 object categories. It is particularly challenging due to the extremely small object scales: the average object size is only 12.7 pixels, with 86\% of instances smaller than 16 pixels and even the largest objects not exceeding 64 pixels.

\subsection{Implementation Details}
\label{subsect4.2:Implement Details}

We train FSDC-DETR for 100 epochs with batch size 64. All input images are normalized and resized to $800\times800$. Our implementation is based on PyTorch 2.5.1, and all experiments conducted on 8$\times$NVIDIA H200 GPUs. We optimize the model using AdamW with the learning rate of $5\times10^{-4}$, momentum 0.9, and weight decay $1.25\times10^{-4}$. 
Following DEIMv2 \cite{DEIMv2}, we employ standard data augmentations such as color jitter and zoom-out, and additionally apply MixUp, Mosaic, and CopyBlend during training. We disable dense O2O \cite{DEIM} after the first 50\% of training epochs, which consistently improves performance. We also turn off all data augmentation in the final two epochs. 
Additional hyper-parameter settings are provided in Supplement Sec.~A.

\subsection{Evaluation Metrics}
\label{subsect4.3:Evaluation Metrics}

To evaluate FSDC-DETR, we report Average Precision (AP) with at most 300 detections per image. Following the standard protocol, AP is computed by averaging results over IoU thresholds from 0.50 to 0.95 with a step size of 0.05. We also report scale-specific AP, i.e., AP$_s$, AP$_m$, and AP$_l$, for small, medium, and large objects, respectively.

\subsection{Main Results}
\label{subsect4.4:Main Results}

\begin{table}[ht]
\centering
\caption{Quantitative comparison on the official test splits of VisDrone-DET2019 and AITODv2 at input resolution $800\times800$.}
\label{tab1:main_results}
\scriptsize
\setlength{\tabcolsep}{1pt}
\renewcommand{\arraystretch}{1.05}

\begin{tabular}{@{}l
                @{\hspace{8pt}}c
                @{\hspace{12pt}}cccccc
                @{\hspace{14pt}}ccccc
                @{}}
\toprule
\multirow{2}{*}{Model} 
& \multirow{2}{*}{Params.} 
& \multicolumn{6}{c}{VisDrone-DET2019} 
& \multicolumn{5}{c}{AITODv2} \\

\cmidrule(lr){3-8}
\cmidrule(lr){9-13}

& & AP & AP$_{50}$ & AP$_{75}$ & AP$_S$ & AP$_M$ & AP$_L$
& AP & AP$_{50}$ & AP$_{75}$ & AP$_S$ & AP$_M$ \\
\midrule

YOLOv11-S\cite{yolo11_ultralytics}      & 9.4    & 19.2 & 32.5 & 19.5 & 10.2 & 28.8 & 42.2  & 15.6
& 33.4 & 12.7 & 14.6 & 31.2 \\
YOLOv12-S\cite{yolov12}                 & 9.3    & 19.5 & 32.9 & 20.2 & 10.5 & 29.5 & 39.8  & 15.1
& 33.1 & 11.9 & 14.2 & 30.7 \\
YOLOv13-S\cite{yolov13}                 & 9.0    & 18.3 & 31.3 & 18.7 &  9.4 & 27.7 & 41.1  & 15.7
& 33.3 & 12.8 & 14.8 & 31.1 \\
FBRT-YOLO-S\cite{FBRT_YOLO_AAAI25}      & 2.9    & 19.5 & 32.9 & 20.2 & 10.1 & 30.0 & 39.9  & 14.3
& 31.1 & 11.2 & 13.2 & 28.8 \\
D-FINE-S\cite{DFine_ICLR2025}           & 10.0   & 23.7 & 41.6 & 23.4 & 14.1 & 33.9 & 47.2  & 24.7
& 54.1 & 18.4 & 23.6 & 41.3 \\
DEIMv2-S\cite{DEIMv2}                   & 10.0   & 23.3 & 41.4 & 23.1 & 13.7 & 33.3 & 47.2  & 23.7
& 53.1 & 18.4 & 22.6 & 40.9 \\
RT-DETRv4-S\cite{RT-DETRv4}             & 10.0   & 23.3 & 41.0 & 23.1 & 13.8 & 33.3 & 43.7  & 23.7
& 53.2 & 18.0 & 22.7 & 39.8 \\
\midrule
YOLOv11-L\cite{yolo11_ultralytics}      & 25.3   & 22.6 & 37.3 & 23.5 & 12.5 & 33.2 & 47.8  & 17.6
& 35.9 & 15.1 & 16.6 & 33.0 \\
YOLOv12-L\cite{yolov12}                 & 26.4   & 21.9 & 36.3 & 22.7 & 12.4 & 32.3 & 43.8  & 17.7
& 36.5 & 15.3 & 16.7 & 33.8 \\
YOLOv13-L\cite{yolov13}                 & 27.6   & 20.6 & 34.4 & 21.1 & 11.7 & 30.4 & 48.7  & 18.0& 36.5 & 15.5 & 17.1 & 32.3 \\
FBRT-YOLO-L\cite{FBRT_YOLO_AAAI25}      & 14.6   & 22.5 & 37.4 & 23.6 & 12.3 & 34.1 & 46.4  & 16.8
& 35.0 & 14.0 & 15.7 & 32.5 \\
D-FINE-L\cite{DFine_ICLR2025}           & 31.0   & 26.1 & 45.6 & 26.4 & 15.8 & 37.5 & 53.6  & 25.1
& 56.2 & 18.8 & 23.9 & 43.2 \\
DEIMv2-L\cite{DEIMv2}                   & 32.0   & 24.7 & 43.3 & 24.6 & 14.2 & 35.4 & 52.7  & 25.7
& 55.7 & 20.6 & 24.4 & 43.7 \\
RT-DETRv4-L\cite{RT-DETRv4}             & 31.0   & 26.4 & 45.7 & 26.7 & 16.1 & 37.3 & 49.3  & 26.8
& 58.3 & 20.6 & 23.0 & 44.5 \\
\midrule
YOLOv11-X\cite{yolo11_ultralytics}      & 56.9   & 21.6 & 35.8 & 22.4 & 12.2 & 31.3 & 47.1  & 18.5
& 36.7 & 16.3 & 17.5 & 34.3 \\
YOLOv12-X\cite{yolov12}                 & 59.1   & 22.2 & 36.8 & 23.3 & 13.0 & 33.1 & 45.0  & 18.4
& 37.2 & 16.7 & 17.3 & 34.3 \\
YOLOv13-X\cite{yolov13}                 & 64.0   & 21.5 & 35.9 & 22.1 & 12.6 & 31.7 & 41.2  & 18.1
& 37.4 & 16.4 & 17.1 & 33.6 \\
FBRT-YOLO-X\cite{FBRT_YOLO_AAAI25}      & 22.8   & 22.9 & 37.6 & 23.7 & 12.7 & 34.6 & 46.8  & 17.2
& 35.7 & 14.4 & 16.2 & 33.0 \\
D-FINE-X\cite{DFine_ICLR2025}           & 62.0   & 26.6 & 46.2 & 26.8 & 15.9 & 38.2 & 56.0  & 26.4
& 58.2 & 20.5 & 25.1 & 45.4 \\
DEIMv2-X\cite{DEIMv2}                   & 50.0   & 26.5 & 45.8 & 26.7 & 16.1 & 37.5 & 52.6  & 25.7
& 59.6 & 22.1 & 26.9 & 48.4 \\
RT-DETRv4-X\cite{RT-DETRv4}             & 62.0   & 27.1 & 46.6 & 27.4 & 16.5 & 38.1 & 53.5  & 27.1
& 59.4 & 21.5 & 26.1 & 45.4 \\
\midrule
VRF-DETR\cite{wenbin2025vrf_detr}       & 13.5   & 23.1 & 40.3 & 22.9 & 14.1 & 32.8 & 38.1  & 16.4
& 39.1 & 12.1 & 16.4 & 27.9 \\
CSFPR-DETR\cite{CSFPR_RTDETR}           & 14.1   & 23.3 & 40.5 & 23.2 & 14.9 & 32.7 & 40.2  & 16.4
& 40.0 & 12.7 & 17.2 & 29.0 \\
UAV-DETR-R18\cite{zhang2025uav_detr}    & 20.0   & 24.2 & 42.0 & 24.1 & 15.0 & 34.2 & 43.8  & 16.7
& 37.9 & 11.8 & 15.7 & 28.1 \\
UAV-DETR-R50\cite{zhang2025uav_detr}    & 42.0   & 25.6 & 43.9 & 25.8 & 15.6 & 36.2 & 45.8  & 17.5
& 37.2 & 13.8 & 16.7 & 35.9 \\
\midrule
\textbf{FSDC-DETR (Ours)} & 40.3  & \textbf{31.1} & \textbf{52.3} & \textbf{31.9} & \textbf{21.0} & \textbf{41.8} & \textbf{58.1}
                  & \textbf{32.3}& \textbf{63.9} & \textbf{28.7} & \textbf{31.3} & \textbf{48.5} \\
\bottomrule
\end{tabular}
\end{table}

Quantitative comparisons on the test splits of VisDrone-DET2019 and AITODv2 are reported in \cref{tab1:main_results}. Consistent improvements over previous state-of-the-art detectors are achieved by the proposed FSDC-DETR, even when compared against their results reported at different model scales.

On VisDrone-DET2019, built upon the DEIMv2-L, FSDC-DETR achieves 31.1 AP, surpassing DEIMv2-L (24.7 AP) by 6.4 points. 
The improvement is more pronounced under AP$_{50}$, where our method reaches 52.3, outperforming DEIMv2-L and RT-DETRv4-L by 9.0 and 6.6 points, respectively. 
Under stricter localization criteria, AP$_{75}$ increases from 24.6 to 31.9 (+7.3), indicating improved BBox precision.
Notably, FSDC-DETR yields substantial gains on small objects. In VisDrone-DET2019, AP$_S$ improves from 14.2 to 21.0, corresponding to a 6.8 point improvement, representing a relative improvement of nearly 48\%. 
These results demonstrate that the proposed frequency-spatial collaborative modeling effectively preserves fine-grained structural details and high-frequency cues during multi-scale aggregation.

The superiority of our design is further validated on AITODv2. FSDC-DETR achieves 32.3 AP and 63.9 AP$_{50}$, surpassing DEIMv2-L by 6.6 AP and 8.2 AP$_{50}$, respectively, and outperforming RT-DETRv4-L by 5.5 AP. 
Moreover, AP$_S$ increases from 24.4 to 31.3 (+6.9) for FSDC-DETR compared to DEIMv2-L, confirming consistent improvements on extremely small objects.
Consistent gains across two challenging benchmarks indicate that FSDC-DETR generalizes robustly across object scales, benefiting not only small instances but also medium and large objects through coherent frequency-spatial interaction.
More quantitative results are provided in Supplement Sec~B, including M-scale comparisons, an enhanced DEIMv2-L baseline with four multi-scale fusion layers, and cross-domain evaluations, further validate the robustness of FSDC-DETR.

\subsection{Ablation Study}
\label{subsec4.5:ablation_study}

\begin{table}[ht]
\centering
\caption{Comprehensive ablation study on the VisDrone-DET2019 test split at input resolution $800\times800$.}
\label{tab2:ablation_results}
\footnotesize
\setlength{\tabcolsep}{6pt}
\begin{tabular}{ccccccccc}
\toprule
 DBFSAF & SFS-FF & FSD-Down & AP & AP$_{50}$ & AP$_{75}$ & AP$_S$ & AP$_M$ & AP$_L$ \\
\midrule
 \ding{55}& \ding{55}&\ding{55}& 24.7 & 43.3 & 24.6 & 14.2 & 35.4 & 52.7 \\
 \ding{51} & \ding{55}& \ding{55}& 29.7 & 49.7 & 30.5 & 19.8 & 40.6 & 55.5 \\
 \ding{55}& \ding{51}& \ding{55}& 25.9& 44.9& 26.0& 15.5& 36.8&53.7\\
 \ding{55}& \ding{55}& \ding{51}& 25.4& 44.2& 25.6& 15.1& 36.4&52.9\\
 \ding{55}& \ding{51}& \ding{51}& 26.3& 45.5& 26.4& 16.4& 37.4&54.5\\
\ding{51}& \ding{55}& \ding{51} & 30.0& 50.5& 30.6& 19.8& 40.8&55.7\\
 \ding{51} & \ding{51} & \ding{55}& 30.3& 50.9& 30.8& 20.0& 41.3& 56.0\\
 \ding{51} & \ding{51} & \ding{51} & \textbf{31.1}& \textbf{52.3} & \textbf{31.9} & \textbf{21.0} & \textbf{41.8} & \textbf{58.1} \\
\bottomrule
\end{tabular}
\end{table}

Ablation results on the VisDrone-DET2019 test split are reported in \cref{tab2:ablation_results}. 
The baseline (DEIMv2-L) achieves 24.7 AP and 43.3 AP$_{50}$. Adding DBFSAF raises performance to 29.7 AP (+5.0), with AP$_{50}$ improving to 49.7 and AP$_{S}$ to 19.8, demonstrating the benefit of frequency-spatial adaptive fusion. 
Incorporating SFS-FF further improves AP to 30.3, yielding consistent gains in AP$_{75}$ and AP$_{S}$. Replacing conventional convolutional downsampling with the proposed FSD-Down results in the best performance of 31.1 AP and 52.3 AP$_{50}$, corresponding to overall improvements of 6.4 AP and 9.0 AP$_{50}$ over the baseline. Notably, AP$_S$ increases from 14.2 to 21.0, highlighting the effectiveness of frequency-preserving downsampling for SOD. These results verify that DBFSAF, SFS-FF, and FSD-Down contribute complementarily to performance gains. 
More ablation studies, module-wise parameter statistics, and frequency-response visualizations are provided in Supplement Sec~C.

\begin{table}[ht]
\centering
\caption{Comprehensive hyperparameter ablation study on the VisDrone-DET2019 test split at input resolution $800\times800$.}
\label{tab3:hyperparam_ablation}
\footnotesize
\setlength{\tabcolsep}{8pt}
\begin{tabular}{ccccccc}\toprule

partial ratio $\gamma$ in DBFSAF& AP & AP$_{50}$ & AP$_{75}$ & AP$_S$ & AP$_M$ & AP$_L$ \\\midrule

 0& 29.7 & 49.7 & 30.5 & 19.8 & 40.6 & 56.5 
\\
 0.375& 30.8 & 51.7 & 31.3 & 20.4 & 41.2 & 57.5 
\\
 \textbf{0.5}& \textbf{31.1} & \textbf{52.3} & \textbf{31.9} & \textbf{21.0} & \textbf{41.8} & \textbf{58.1} 
\\
 0.625& 30.9 & 51.8 & 31.4 & 20.5 & 41.2 & 57.7 
\\

 1& 30.0 & 50.4 & 30.7 & 20.0 & 41.0 &56.4 
\\ \bottomrule 
\end{tabular}

\end{table}

Performance improves as the partial ratio $\gamma$ in DBFSAF, as defined in Eq.~\ref{eq:1}, increases from 0 to 0.5, achieving the best results at $\gamma=0.5$ with 31.1 AP and 21.0 AP$_S$, as shown in \cref{tab3:hyperparam_ablation}. 
A further increase in $\gamma$ slightly degrades performance, indicating that excessive refinement may introduce redundant interactions. The noticeable drop in $\gamma=0$ further confirms the effectiveness of DBFSAF.

\subsection{Visualization Analysis}
\label{subsec4.6:Visualization Analysis}

\begin{figure}[!ht]
    \centering
    \includegraphics[width=0.95\linewidth]{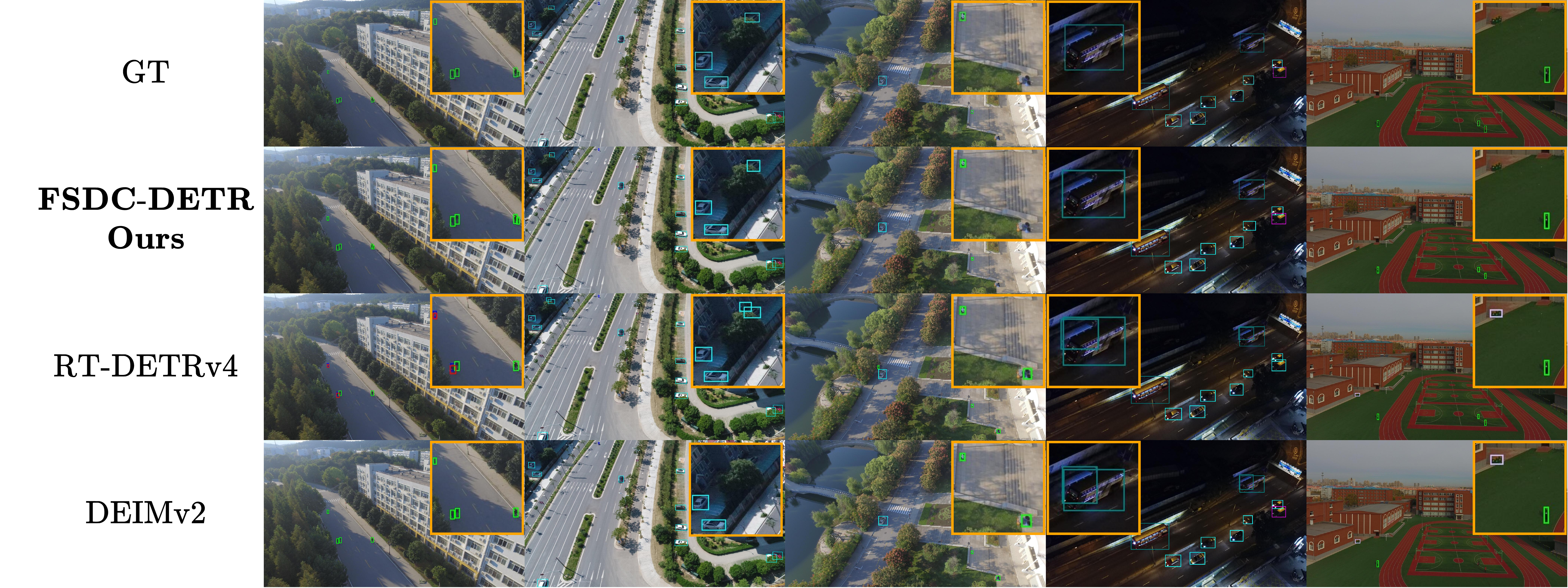}
    \caption{Visualization comparison of small object detection results on the VisDrone-DET2019 test split. \textcolor{orange}{Orange boxes} indicate zoomed-in details.}
    \label{fig6:visdrone-vision}
\end{figure}

\begin{figure}[ht]
    \centering
    \includegraphics[width=0.95\linewidth]{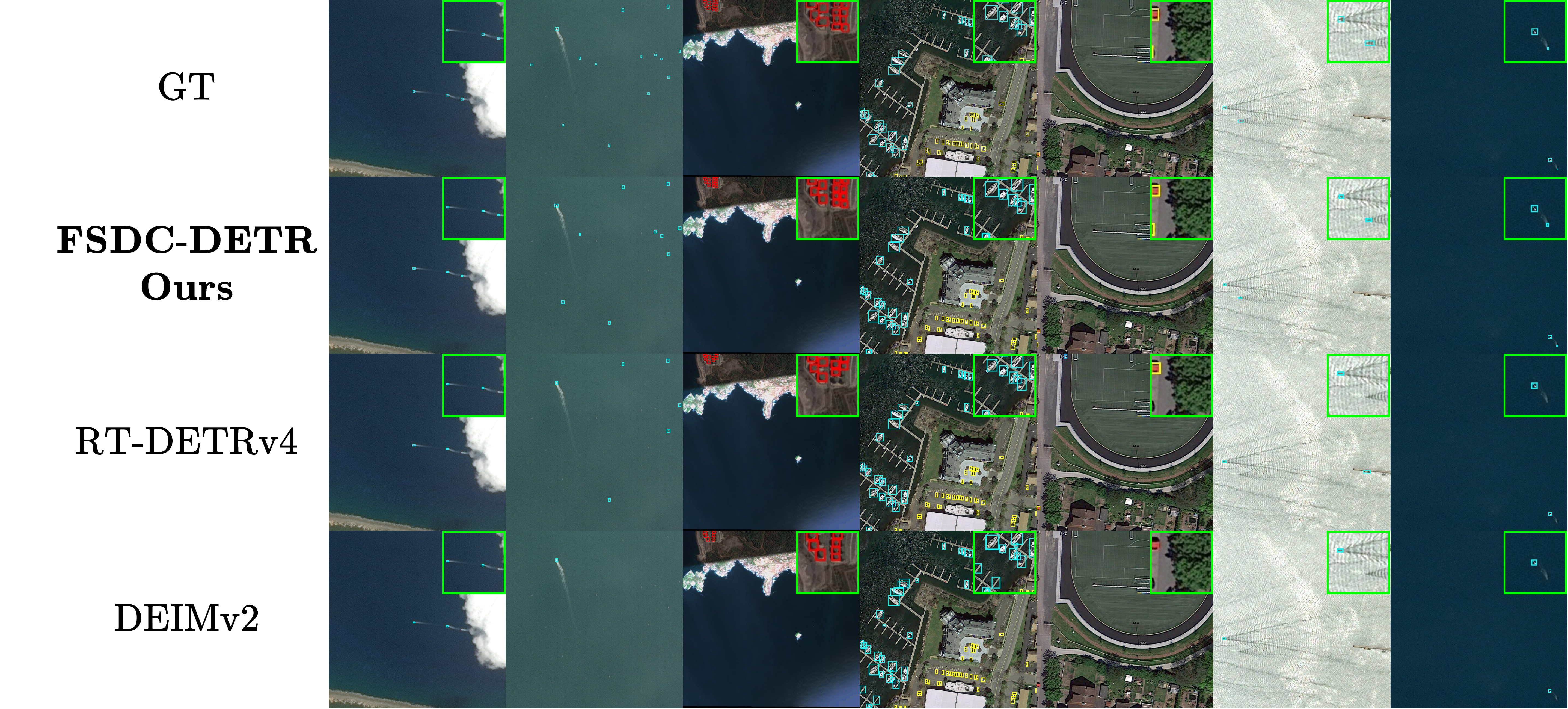}
    \caption{Visualization comparison of small object detection results on the AITODv2 test split. \textcolor{green}{Green boxes} indicate zoomed-in details.}
    \label{fig7:aitodv2_vision}
\end{figure}

Visualization comparisons between FSDC-DETR and competing models on the VisDrone-DET2019 and AITODv2 test splits are presented in Fig.~\ref{fig6:visdrone-vision} and Fig.~\ref{fig7:aitodv2_vision}. 
Benefiting from the integration of DBFSAF, SFS-FF, and FSD-Down, FSDC-DETR demonstrates superior capability in SOD. 

On the VisDrone-DET2019 test split, FSDC-DETR exhibits more precise localization of small objects, as shown in the first two columns of Fig.~\ref{fig6:visdrone-vision}, where object boundaries are clearer and object regions are more accurately aligned with ground truth (GT). As illustrated in columns 3-5, our model maintains robust detection under cluttered and complex backgrounds while effectively reducing false positives compared to competing methods.

On the AITODv2 test split, as shown in Fig.~\ref{fig7:aitodv2_vision}, FSDC-DETR achieves consistently more precise localization of tiny objects. In columns 1-2 and 5-7, small instances are more precisely detected and exhibit better alignment with the GT. In columns 3-4, the proposed method effectively suppresses false positives and maintains stable detection performance under challenging conditions, further demonstrating the robustness of the frequency-spatial collaborative modeling.
Supplement Sec.~D provides additional heatmap visualizations and failure case analyses of FSDC-DETR.

\section{Conclusion}
\label{sec5:conclusion}

In this paper, we present FSDC-DETR for the precise SOD. Unlike conventional detectors that implicitly entangle spatial aggregation with frequency attenuation, FSDC-DETR explicitly constructs, propagates, and preserves frequency-aware representations throughout the detection pipeline.
The proposed DBFSAF module constructs frequency-aware representations by mitigating frequency aliasing and truncation in ViT-CNN backbones through adaptive high-frequency preservation and frequency-spatial collaborative refinement.
Building upon these refined representations, the SFS-FF module propagates frequency-enriched features across scales via structure-aware frequency-spatial collaborative aggregation.
Furthermore, the FSD-Down preserves frequency integrity during scale transitions while maintaining spatial consistency.
By systematically addressing representation construction, cross-scale propagation, and scale-transition preservation from a unified frequency-spatial perspective, FSDC-DETR significantly improves SOD performance. 
Experimental results on two challenging benchmarks confirm state-of-the-art performance, demonstrating excellent detection capabilities for medium and large objects, with particularly significant improvements in SOD.


\section*{Acknowledgements}

This work was supported in part by the Changzhou Longcheng Talent Program under Grant No.~CQ20250117 and in part by the National Natural Science Foundation of China under Grant No.~62377011.

%
%
\bibliographystyle{splncs04}
\bibliography{main}

@String(CVPR  = {IEEE Conf. Comput. Vis. Pattern Recog.})

@String(ICCV  = {Int. Conf. Comput. Vis.})

@String(ECCV  = {Eur. Conf. Comput. Vis.})

@String(CVPRW = {IEEE Conf. Comput. Vis. Pattern Recog. Worksh.})

@String(AAAI  = {AAAI})

@String(ICPR  = {Int. Conf. Pattern Recog.})

@inproceedings{VisDrone-DET2019,
  author    = {Du, Dawei and Zhu, Pengfei and Wen, Longyin and Bian, Xiao and Lin, Haibin and Hu, Qinghua and others},
  booktitle = {2019 IEEE/CVF International Conference on Computer Vision Workshop (ICCVW)},
  title     = {VisDrone-DET2019: The Vision Meets Drone Object Detection in Image Challenge Results},
  year      = {2019},
  pages     = {213--226},
  doi       = {10.1109/ICCVW.2019.00030}
}

@inproceedings{AITODv1,
  author={Wang, Jinwang and Yang, Wen and Guo, Haowen and Zhang, Ruixiang and Xia, Gui-Song},
  booktitle={2020 25th International Conference on Pattern Recognition (ICPR)}, 
  title={Tiny Object Detection in Aerial Images}, 
  year={2021},
  volume={},
  number={},
  pages={3791-3798},
  doi={10.1109/ICPR48806.2021.9413340}}

@article{AITODv2,
title = {Detecting tiny objects in aerial images: A normalized Wasserstein distance and a new benchmark},
journal = {ISPRS Journal of Photogrammetry and Remote Sensing},
volume = {190},
pages = {79-93},
year = {2022},
issn = {0924-2716},
doi = {https://doi.org/10.1016/j.isprsjprs.2022.06.002},
url = {https://www.sciencedirect.com/science/article/pii/S0924271622001599},
author = {Chang Xu and Jinwang Wang and Wen Yang and Huai Yu and Lei Yu and Gui-Song Xia}
}

@inproceedings{RT-DETRv1,
  author={Zhao, Yian and Lv, Wenyu and Xu, Shangliang and Wei, Jinman and Wang, Guanzhong and Dang, Qingqing and others},
  booktitle={2024 IEEE/CVF Conference on Computer Vision and Pattern Recognition (CVPR)}, 
  title={DETRs Beat YOLOs on Real-time Object Detection}, 
  year={2024},
  volume={},
  number={},
  pages={16965-16974},
  doi={10.1109/CVPR52733.2024.01605}}

@article{RT-DETRv2,
  title   = {Rt-detrv2: Improved baseline with bag-of-freebies for real-time detection transformer},
  author  = {Lv, Wenyu and Zhao, Yian and Chang, Qinyao and Huang, Kui and Wang, Guanzhong and Liu, Yi},
  journal = {arXiv preprint arXiv:2407.17140},
  year    = {2024}
}

@inproceedings{RT-DETRv3,
  author={Wang, Shuo and Xia, Chunlong and Lv, Feng and Shi, Yifeng},
  booktitle={2025 IEEE/CVF Winter Conference on Applications of Computer Vision (WACV)}, 
  title={RT-DETRv3: Real-Time End-to-End Object Detection with Hierarchical Dense Positive Supervision}, 
  year={2025},
  volume={},
  number={},
  pages={1628-1636},
  doi={10.1109/WACV61041.2025.00166}}

@article{RT-DETRv4,
  title={RT-DETRv4: Painlessly Furthering Real-Time Object Detection with Vision Foundation Models},
  author={Zijun Liao and Yian Zhao and Xin Shan and Yu Yan and Chang Liu and Lei Lu and Xiangyang Ji and Jie Chen},
  journal={arXiv preprint arXiv:2510.25257},
  year={2025}
}

@inproceedings{DFine_ICLR2025,
  title={D-FINE: Redefine regression task of DETRs as fine-grained distribution refinement},
  author={Peng, Yansong and Li, Hebei and Wu, Peixi and Zhang, Yueyi and Sun, Xiaoyan and Wu, Feng},
  booktitle={International Conference on Learning Representations},
  volume={2025},
  pages={44015--44031},
  year={2025}
}

@inproceedings{DEIM,
  author={Huang, Shihua and Lu, Zhichao and Cun, Xiaodong and Yu, Yongjun and Zhou, Xiao and Shen, Xi},
  booktitle={2025 IEEE/CVF Conference on Computer Vision and Pattern Recognition (CVPR)}, 
  title={DEIM: DETR with Improved Matching for Fast Convergence}, 
  year={2025},
  volume={},
  number={},
  pages={15162-15171},
  doi={10.1109/CVPR52734.2025.01412}}

@article{DEIMv2,
  title={Real-time object detection meets DINOv3},
  author={Huang, Shihua and Hou, Yongjie and Liu, Longfei and Yu, Xuanlong and Shen, Xi},
  journal={arXiv preprint arXiv:2509.20787},
  year={2025}
}

@inproceedings{Chenlinwei_FDConv_CVPR25,
  author={Chen, Linwei and Gu, Lin and Li, Liang and Yan, Chenggang and Fu, Ying},
  booktitle={2025 IEEE/CVF Conference on Computer Vision and Pattern Recognition (CVPR)}, 
  title={Frequency Dynamic Convolution for Dense Image Prediction}, 
  year={2025},
  volume={},
  number={},
  pages={30178-30188},
  doi={10.1109/CVPR52734.2025.02809}}

@inproceedings{Chen_2025_ICCV,
  author={Chen, Linwei and Gu, Lin and Fu, Ying},
  booktitle={2025 IEEE/CVF International Conference on Computer Vision (ICCV)}, 
  title={Frequency-Dynamic Attention Modulation for Dense Prediction}, 
  year={2025},
  volume={},
  number={},
  pages={22620-22632},
  doi={10.1109/ICCV51701.2025.02100}}

@article{shu2026waveformer,
  title={WaveFormer: Frequency-Time Decoupled Vision Modeling with Wave Equation},
  author={Shu, Zishan and Wu, Juntong and Yan, Wei and Liu, Xudong and Zhang, Hongyu and Liu, Chang and Mao, Youdong and Chen, Jie},
  journal={arXiv preprint arXiv:2601.08602},
  year={2026}
}

@article{wang2022antioversmooth,
  title={Anti-oversmoothing in deep vision transformers via the fourier domain analysis: From theory to practice},
  author={Wang, Peihao and Zheng, Wenqing and Chen, Tianlong and Wang, Zhangyang},
  journal={arXiv preprint arXiv:2203.05962},
  year={2022}
}

@inproceedings{tatsunami2024dynamic-filter,
    author = {Tatsunami, Yuki and Taki, Masato},
    title = {FFT-based dynamic token mixer for vision},
    year = {2024},
    isbn = {978-1-57735-887-9},
    publisher = {AAAI Press},
    url = {https://doi.org/10.1609/aaai.v38i14.29457},
    doi = {10.1609/aaai.v38i14.29457},
    booktitle = {Proceedings of the Thirty-Eighth AAAI Conference on Artificial Intelligence and Thirty-Sixth Conference on Innovative Applications of Artificial Intelligence and Fourteenth Symposium on Educational Advances in Artificial Intelligence},
    articleno = {1709},
    numpages = {9},
    series = {AAAI'24/IAAI'24/EAAI'24}
}

@inproceedings{caron2021dinov1,
    author={Caron, Mathilde and Touvron, Hugo and Misra, Ishan and Jegou, Hervé and Mairal, Julien and Bojanowski, Piotr and Joulin, Armand},
    booktitle={2021 IEEE/CVF International Conference on Computer Vision (ICCV)}, 
    title={Emerging Properties in Self-Supervised Vision Transformers}, 
    year={2021},
    volume={},
    number={},
    pages={9630-9640},
    doi={10.1109/ICCV48922.2021.00951}
}

@article{oquab2023dinov2,
  title={Dinov2: Learning robust visual features without supervision},
  author={Oquab, Maxime and Darcet, Timoth{\'e}e and Moutakanni, Th{\'e}o and Vo, Huy and Szafraniec, Marc and Khalidov, Vasil and Fernandez, Pierre and Haziza, Daniel and Massa, Francisco and El-Nouby, Alaaeldin and others},
  journal={arXiv preprint arXiv:2304.07193},
  year={2023}
}

@article{simeoni2025dinov3,
  title={Dinov3},
  author={Sim{\'e}oni, Oriane and Vo, Huy V and Seitzer, Maximilian and Baldassarre, Federico and Oquab, Maxime and Jose, Cijo and others},
  journal={arXiv preprint arXiv:2508.10104},
  year={2025}
}

@article{carion2025sam3segmentconcepts,
  title={Sam 3: Segment anything with concepts},
  author={Carion, Nicolas and Gustafson, Laura and Hu, Yuan-Ting and Debnath, Shoubhik and Hu, Ronghang and Suris, Didac and Ryali, Chaitanya and Alwala, Kalyan Vasudev and Khedr, Haitham and Huang, Andrew and others},
  journal={arXiv preprint arXiv:2511.16719},
  year={2025}
}

@article{ranzinger2026cradiov4techreport,
  title={C-RADIOv4 (Tech Report)},
  author={Ranzinger, Mike and Heinrich, Greg and McCarthy, Collin and Kautz, Jan and Tao, Andrew and Catanzaro, Bryan and Molchanov, Pavlo},
  journal={arXiv preprint arXiv:2601.17237},
  year={2026}
}

@inproceedings{Wang_CNN_high_freq_cvpr20,
  author={Wang, Haohan and Wu, Xindi and Huang, Zeyi and Xing, Eric P.},
  booktitle={2020 IEEE/CVF Conference on Computer Vision and Pattern Recognition (CVPR)}, 
  title={High-Frequency Component Helps Explain the Generalization of Convolutional Neural Networks}, 
  year={2020},
  volume={},
  number={},
  pages={8681-8691},
  doi={10.1109/CVPR42600.2020.00871}
}

@article{cui2021hgnetv2,
  title={Beyond self-supervision: A simple yet effective network distillation alternative to improve backbones},
  author={Cui, Cheng and Guo, Ruoyu and Du, Yuning and He, Dongliang and Li, Fu and Wu, Zewu and Liu, Qiwen and Wen, Shilei and Huang, Jizhou and Hu, Xiaoguang and others},
  journal={arXiv preprint arXiv:2103.05959},
  year={2021}
}

@inproceedings{Li_SFSConv_CVPR25,
  author    = {Li, Ke and Wang, Di and Hu, Zhangyuan and Zhu, Wenxuan and Li, Shaofeng and Wang, Quan},
  title     = {Unleashing Channel Potential: Space-Frequency Selection Convolution for SAR Object Detection},
  booktitle = {2024 IEEE/CVF Conference on Computer Vision and Pattern Recognition (CVPR)},
  year      = {2024},
  pages     = {17323--17332},
  doi       = {10.1109/CVPR52733.2024.01640},
}

@article{huang2025partialchannelnetworkcompute_AAAI26,
  title={Partial channel network: Compute fewer, perform better},
  author={Huang, Haiduo and Xia, Tian and Ren, Pengju and others},
  journal={arXiv preprint arXiv:2502.01303},
  year={2025}
}

@inproceedings{9522696,
  author={Abello, Antonio A. and Hirata, Roberto and Wang, Zhangyang},
  booktitle={2021 IEEE/CVF Conference on Computer Vision and Pattern Recognition Workshops (CVPRW)}, 
  title={Dissecting the High-Frequency Bias in Convolutional Neural Networks}, 
  year={2021},
  volume={},
  number={},
  pages={863--871},
  doi={10.1109/CVPRW53098.2021.00096},
}

@inproceedings{paul2022vision,
  title     = {Vision transformers are robust learners},
  author    = {Paul, Sayak and Chen, Pin-Yu},
  booktitle = {Proceedings of the AAAI Conference on Artificial Intelligence (AAAI)},
  volume    = {36},
  pages     = {2071--2081},
  year      = {2022}
}

@inproceedings{rahman2025mkunet,
  author={Rahman, Md Mostafijur and Marculescu, Radu},
  booktitle={2025 IEEE/CVF International Conference on Computer Vision Workshops (ICCVW)}, 
  title={MK-UNet: Multi-Kernel Lightweight CNN for Medical Image Segmentation}, 
  year={2025},
  volume={},
  number={},
  pages={1053-1062},
  doi={10.1109/ICCVW69036.2025.00114}
}

@inproceedings{sun2024FESL,
    author="Sun, Yanguang
    and Xu, Chunyan
    and Yang, Jian
    and Xuan, Hanyu
    and Luo, Lei",
    editor="Leonardis, Ale{\v{s}}
    and Ricci, Elisa
    and Roth, Stefan
    and Russakovsky, Olga
    and Sattler, Torsten
    and Varol, G{\"u}l",
    title="Frequency-Spatial Entanglement Learning for Camouflaged Object Detection",
    booktitle="Computer Vision -- ECCV 2024",
    year="2025",
    publisher="Springer Nature Switzerland",
    address="Cham",
    pages="343--360",
    isbn="978-3-031-72658-3"
}

@inproceedings{zhang2025decoding_aaai26,
  title={Decoding with structured awareness: integrating directional, frequency-spatial, and structural attention for medical image segmentation},
  author={Zhang, Fan and Gu, Zhiwei and Wang, Hua},
  booktitle={Proceedings of the AAAI Conference on Artificial Intelligence},
  volume={40},
  pages={12421--12429},
  year={2026}
}

@inproceedings{liu2025lidar_ACM_MM25,
    author = {Liu, Hui and Jia, Chen and Shi, Fan and Cheng, Xu and Shi, Mengfei and Xie, Xia and others},
    title = {LIDAR: Lightweight Adaptive Cue-Aware Fusion Vision Mamba for Multimodal Segmentation of Structural Cracks},
    year = {2025},
    isbn = {9798400720352},
    publisher = {Association for Computing Machinery},
    address = {New York, NY, USA},
    url = {https://doi.org/10.1145/3746027.3755452},
    doi = {10.1145/3746027.3755452},
    booktitle = {Proceedings of the 33rd ACM International Conference on Multimedia},
    pages = {1832–1841},
    numpages = {10},
    location = {Dublin, Ireland},
    series = {MM '25}
}

@article{hwd_xu_pr23,
    title = {Haar wavelet downsampling: A simple but effective downsampling module for semantic segmentation},
    journal = {Pattern Recognition},
    volume = {143},
    pages = {109819},
    year = {2023},
    issn = {0031-3203},
    doi = {https://doi.org/10.1016/j.patcog.2023.109819},
    url = {https://www.sciencedirect.com/science/article/pii/S0031320323005174},
    author = {Guoping Xu and Wentao Liao and Xuan Zhang and Chang Li and Xinwei He and Xinglong Wu}
}

@inproceedings{zhang2025uav_detr,
  author={Zhang, Huaxiang and Zhang, Hao and Liu, Kai and Gan, Zhongxue and Zhu, Guo-Niu},
  booktitle={2025 IEEE/RSJ International Conference on Intelligent Robots and Systems (IROS)}, 
  title={UAV-DETR: Efficient End-to-End Object Detection for Unmanned Aerial Vehicle Imagery}, 
  year={2025},
  volume={},
  number={},
  pages={15143-15149},
  doi={10.1109/IROS60139.2025.11246176}}

@inproceedings{FBRT_YOLO_AAAI25,
  title={Fbrt-yolo: Faster and better for real-time aerial image detection},
  author={Xiao, Yao and Xu, Tingfa and Xin, Yu and Li, Jianan},
  booktitle={Proceedings of the AAAI Conference on Artificial Intelligence},
  volume={39},
  pages={8673--8681},
  year={2025}
}

@inproceedings{fast_rcnn,
  author={Girshick, Ross},
  booktitle={2015 IEEE International Conference on Computer Vision (ICCV)}, 
  title={Fast R-CNN}, 
  year={2015},
  volume={},
  number={},
  pages={1440-1448},
  doi={10.1109/ICCV.2015.169}}

@ARTICLE{faster_rcnn,
  author={Ren, Shaoqing and He, Kaiming and Girshick, Ross and Sun, Jian},
  journal={IEEE Transactions on Pattern Analysis and Machine Intelligence}, 
  title={Faster R-CNN: Towards Real-Time Object Detection with Region Proposal Networks}, 
  year={2017},
  volume={39},
  number={6},
  pages={1137-1149},
  doi={10.1109/TPAMI.2016.2577031}}

@article{yolov5,
  title={What is YOLOv5: A deep look into the internal features of the popular object detector},
  author={Khanam, Rahima and Hussain, Muhammad},
  journal={arXiv preprint arXiv:2407.20892},
  year={2024}
}

@inproceedings{yolov8,
  author={Varghese, Rejin and M., Sambath},
  booktitle={2024 International Conference on Advances in Data Engineering and Intelligent Computing Systems (ADICS)}, 
  title={YOLOv8: A Novel Object Detection Algorithm with Enhanced Performance and Robustness}, 
  year={2024},
  volume={},
  number={},
  pages={1-6},
  doi={10.1109/ADICS58448.2024.10533619}
}

@inproceedings{yolov9,
    author="Wang, Chien-Yao
    and Yeh, I-Hau
    and Mark Liao, Hong-Yuan",
    editor="Leonardis, Ale{\v{s}}
    and Ricci, Elisa
    and Roth, Stefan
    and Russakovsky, Olga
    and Sattler, Torsten
    and Varol, G{\"u}l",
    title="YOLOv9: Learning What You Want to Learn Using Programmable Gradient Information",
    booktitle="Computer Vision -- ECCV 2024",
    year="2025",
    publisher="Springer Nature Switzerland",
    address="Cham",
    pages="1--21",
    isbn="978-3-031-72751-1"
}

@inproceedings{yolov10,
    author = {Wang, Ao and Chen, Hui and Liu, Lihao and Chen, Kai and Lin, Zijia and Han, Jungong and others},
    title = {YOLOv10: real-time end-to-end object detection},
    year = {2024},
    isbn = {9798331314385},
    publisher = {Curran Associates Inc.},
    address = {Red Hook, NY, USA},
    booktitle = {Proceedings of the 38th International Conference on Neural Information Processing Systems},
    articleno = {3429},
    numpages = {28},
    location = {Vancouver, BC, Canada},
    series = {NIPS '24}
}

@article{yolo11_ultralytics,
  title={Yolov11: An overview of the key architectural enhancements},
  author={Khanam, Rahima and Hussain, Muhammad},
  journal={arXiv preprint arXiv:2410.17725},
  year={2024}
}

@inproceedings{yolov12,
 author = {Tian, Yunjie and Ye, Qixiang and Doermann, David},
 booktitle = {Advances in Neural Information Processing Systems},
 editor = {D. Belgrave and C. Zhang and H. Lin and R. Pascanu and P. Koniusz and M. Ghassemi and N. Chen},
 pages = {78433--78457},
 publisher = {Curran Associates, Inc.},
 title = {YOLOv12: Attention-Centric Real-Time Object Detectors},
 url = {https://proceedings.neurips.cc/paper\_files/paper/2025/file/7103444259031cc58051f8c9a4868533-Paper-Conference.pdf},
 volume = {38},
 year = {2025}
}

@article{yolov13,
  title={Yolov13: Real-time object detection with hypergraph-enhanced adaptive visual perception},
  author={Lei, Mengqi and Li, Siqi and Wu, Yihong and Hu, Han and Zhou, You and Zheng, Xinhu and Ding, Guiguang and Du, Shaoyi and Wu, Zongze and Gao, Yue},
  journal={arXiv preprint arXiv:2506.17733},
  year={2025}
}

@article{yolo26_ultralytics,
  title={YOLO26: key architectural enhancements and performance benchmarking for real-time object detection},
  author={Sapkota, Ranjan and Cheppally, Rahul Harsha and Sharda, Ajay and Karkee, Manoj},
  journal={arXiv preprint arXiv:2509.25164},
  year={2025}
}

@inproceedings{mamba_yolo,
    author = {Wang, Zeyu and Li, Chen and Xu, Huiying and Zhu, Xinzhong and Li, Hongbo},
    title = {Mamba YOLO: a simple baseline for object detection with state space model},
    year = {2025},
    isbn = {978-1-57735-897-8},
    publisher = {AAAI Press},
    url = {https://doi.org/10.1609/aaai.v39i8.32885},
    doi = {10.1609/aaai.v39i8.32885},
    booktitle = {Proceedings of the Thirty-Ninth AAAI Conference on Artificial Intelligence and Thirty-Seventh Conference on Innovative Applications of Artificial Intelligence and Fifteenth Symposium on Educational Advances in Artificial Intelligence},
    articleno = {912},
    numpages = {9},
    series = {AAAI'25/IAAI'25/EAAI'25}
}

@inproceedings{DETR_ECCV20,
    author="Carion, Nicolas
    and Massa, Francisco
    and Synnaeve, Gabriel
    and Usunier, Nicolas
    and Kirillov, Alexander
    and Zagoruyko, Sergey",
    editor="Vedaldi, Andrea
    and Bischof, Horst
    and Brox, Thomas
    and Frahm, Jan-Michael",
    title="End-to-End Object Detection with Transformers",
    booktitle="Computer Vision -- ECCV 2020",
    year="2020",
    publisher="Springer International Publishing",
    address="Cham",
    pages="213--229",
    isbn="978-3-030-58452-8"
}

@article{wenbin2025vrf_detr,
  title={An efficient aerial image detection with variable receptive fields},
  author={Wenbin, Liu},
  journal={arXiv preprint arXiv:2504.15165},
  year={2025}
}

@ARTICLE{CSFPR_RTDETR,
  author={Hu, Lei and Yuan, Jiwen and Cheng, Bailiang and Xu, Qizhi},
  journal={IEEE Transactions on Geoscience and Remote Sensing}, 
  title={CSFPR-RTDETR: Real-Time Small Object Detection Network for UAV Images Based on Cross-Spatial-Frequency Domain and Position Relation}, 
  year={2025},
  volume={63},
  number={},
  pages={1-19},
  doi={10.1109/TGRS.2025.3601828}}

@article{RF_DETR,
  title={RF-DETR: neural architecture search for real-time detection transformers},
  author={Robinson, Isaac and Robicheaux, Peter and Popov, Matvei and Ramanan, Deva and Peri, Neehar},
  journal={arXiv preprint arXiv:2511.09554},
  year={2025}
}

@book{gonzalez2009digital,
  title     = {Digital image processing},
  author    = {Gonzalez, Rafael C},
  year      = {2009},
  publisher = {Pearson Education India}
}

@book{pitas2000digital,
  title     = {Digital image processing algorithms and applications},
  author    = {Pitas, Ioannis},
  year      = {2000},
  publisher = {John Wiley \& Sons}
}

@inproceedings{yin2019fourier,
    author = {Yin, Dong and Lopes, Raphael Gontijo and Shlens, Jonathon and Cubuk, Ekin D. and Gilmer, Justin},
    title = {A fourier perspective on model robustness in computer vision},
    year = {2019},
    publisher = {Curran Associates Inc.},
    address = {Red Hook, NY, USA},
    booktitle = {Proceedings of the 33rd International Conference on Neural Information Processing Systems},
    articleno = {1189},
    numpages = {11}
}

@inproceedings{huang2023adaptive,
  author={Huang, Zhipeng and Zhang, Zhizheng and Lan, Cuiling and Zha, Zheng-Jun and Lu, Yan and Guo, Baining},
  booktitle={2023 IEEE/CVF International Conference on Computer Vision (ICCV)}, 
  title={Adaptive Frequency Filters As Efficient Global Token Mixers}, 
  year={2023},
  volume={},
  number={},
  pages={6026-6036},
  doi={10.1109/ICCV51070.2023.00556}
}

@inproceedings{shi2025hs_fpn,
    author = {Shi, Zican and Hu, Jing and Ren, Jie and Ye, Hengkang and Yuan, Xuyang and Ouyang, Yan and He, Jia and Ji, Bo and Guo, Junyu},
    title = {HS-FPN: high frequency and spatial perception FPN for tiny object detection},
    year = {2025},
    isbn = {978-1-57735-897-8},
    publisher = {AAAI Press},
    url = {https://doi.org/10.1609/aaai.v39i7.32740},
    doi = {10.1609/aaai.v39i7.32740},
    booktitle = {Proceedings of the Thirty-Ninth AAAI Conference on Artificial Intelligence and Thirty-Seventh Conference on Innovative Applications of Artificial Intelligence and Fifteenth Symposium on Educational Advances in Artificial Intelligence},
    articleno = {767},
    numpages = {9},
    series = {AAAI'25/IAAI'25/EAAI'25}
}

@inproceedings{Li_2025_ICCV_waveseg,
  author={Li, Jiajia and Wu, Huisi and Qin, Jing},
  booktitle={2025 IEEE/CVF International Conference on Computer Vision (ICCV)}, 
  title={WeaveSeg: Iterative Contrast-weaving and Spectral Feature-refining for Nuclei Instance Segmentation}, 
  year={2025},
  volume={},
  number={},
  pages={21984-21993},
  doi={10.1109/ICCV51701.2025.02041}}

@ARTICLE{TIP25_Spatial_Frequency_mamba,
  author={Sun, Hui and Lv, Long and Zhang, Pingping and Tang, Tongdan and Tian, Feng and Sun, Weibing and Lu, Huchuan},
  journal={IEEE Transactions on Image Processing}, 
  title={Spatial-Frequency Enhanced Mamba for Multi-Modal Image Fusion}, 
  year={2025},
  volume={34},
  number={},
  pages={7684-7696},
  doi={10.1109/TIP.2025.3632221}
}

@inproceedings{li2025noise_CVPR25,
  author={Li, Hesong and Wu, Ziqi and Shao, Ruiwen and Zhang, Tao and Fu, Ying},
  booktitle={2025 IEEE/CVF Conference on Computer Vision and Pattern Recognition (CVPR)}, 
  title={Noise Calibration and Spatial-Frequency Interactive Network for STEM Image Enhancement}, 
  year={2025},
  volume={},
  number={},
  pages={21287-21296},
  doi={10.1109/CVPR52734.2025.01983}
}

@inproceedings{Yan_2025_CVPR_WPFormer,
  author={Yan, Feng and Jiang, Xiaoheng and Lu, Yang and Cao, Jiale and Chen, Dong and Xu, Mingliang},
  booktitle={2025 IEEE/CVF Conference on Computer Vision and Pattern Recognition (CVPR)}, 
  title={Wavelet and Prototype Augmented Query-based Transformer for Pixel-level Surface Defect Detection}, 
  year={2025},
  volume={},
  number={},
  pages={23860-23869},
  doi={10.1109/CVPR52734.2025.02222}
}

@article{park2022how,
  title={How do vision transformers work?},
  author={Park, Namuk and Kim, Songkuk},
  journal={arXiv preprint arXiv:2202.06709},
  year={2022}
}

@inproceedings{Zhou_2017_CVPR,
  author={Zhou, Bolei and Zhao, Hang and Puig, Xavier and Fidler, Sanja and Barriuso, Adela and Torralba, Antonio},
  booktitle={2017 IEEE Conference on Computer Vision and Pattern Recognition (CVPR)}, 
  title={Scene Parsing through ADE20K Dataset}, 
  year={2017},
  volume={},
  number={},
  pages={5122-5130},
  doi={10.1109/CVPR.2017.544}
}

@inproceedings{qin2021fcanet,
  author={Qin, Zequn and Zhang, Pengyi and Wu, Fei and Li, Xi},
  booktitle={2021 IEEE/CVF International Conference on Computer Vision (ICCV)}, 
  title={FcaNet: Frequency Channel Attention Networks}, 
  year={2021},
  volume={},
  number={},
  pages={763-772},
  doi={10.1109/ICCV48922.2021.00082}}

@ARTICLE{FreqFusion,
  author={Chen, Linwei and Fu, Ying and Gu, Lin and Yan, Chenggang and Harada, Tatsuya and Huang, Gao},
  journal={IEEE Transactions on Pattern Analysis and Machine Intelligence}, 
  title={Frequency-Aware Feature Fusion for Dense Image Prediction}, 
  year={2024},
  volume={46},
  number={12},
  pages={10763-10780},
  doi={10.1109/TPAMI.2024.3449959}}

@inproceedings{copy_paste,
  author={Ghiasi, Golnaz and Cui, Yin and Srinivas, Aravind and Qian, Rui and Lin, Tsung-Yi and Cubuk, Ekin D. and Le, Quoc V. and Zoph, Barret},
  booktitle={2021 IEEE/CVF Conference on Computer Vision and Pattern Recognition (CVPR)}, 
  title={Simple Copy-Paste is a Strong Data Augmentation Method for Instance Segmentation}, 
  year={2021},
  volume={},
  number={},
  pages={2917-2927},
  doi={10.1109/CVPR46437.2021.00294}}

@ARTICLE{SFM_BIT_Chenlinwei,
  author={Chen, Linwei and Fu, Ying and Gu, Lin and Zheng, Dezhi and Dai, Jifeng},
  journal={IEEE Transactions on Pattern Analysis and Machine Intelligence}, 
  title={Spatial Frequency Modulation for Semantic Segmentation}, 
  year={2025},
  volume={47},
  number={11},
  pages={9767-9784},
  doi={10.1109/TPAMI.2025.3592621}}

@inproceedings{FADC_BIT_Chenlinwei,
  author={Chen, Linwei and Gu, Lin and Zheng, Dezhi and Fu, Ying},
  booktitle={2024 IEEE/CVF Conference on Computer Vision and Pattern Recognition (CVPR)}, 
  title={Frequency-Adaptive Dilated Convolution for Semantic Segmentation}, 
  year={2024},
  volume={},
  number={},
  pages={3414-3425},
  doi={10.1109/CVPR52733.2024.00328}}

@inproceedings{FCENet_CVPR25,
  author={Wang, Yuchen and Wang, Hongyuan and Wang, Lizhi and Wang, Xin and Zhu, Lin and Lu, Wanxuan and others},
  booktitle={2025 IEEE/CVF Conference on Computer Vision and Pattern Recognition (CVPR)}, 
  title={Complementary Advantages: Exploiting Cross-Field Frequency Correlation for NIR-Assisted Image Denoising}, 
  year={2025},
  volume={},
  number={},
  pages={12679-12689},
  doi={10.1109/CVPR52734.2025.01183}}

@inproceedings{zhang2025mrdetr,
  author={Zhang, Chang-Bin and Zhong, Yujie and Han, Kai},
  booktitle={2025 IEEE/CVF Conference on Computer Vision and Pattern Recognition (CVPR)}, 
  title={Mr. DETR: Instructive Multi-Route Training for Detection Transformers}, 
  year={2025},
  volume={},
  number={},
  pages={9933-9943},
  doi={10.1109/CVPR52734.2025.00928}}

@article{hua2025survey,
  title={A survey of small object detection based on deep learning in aerial images},
  author={Hua, Wei and Chen, Qili},
  journal={Artificial Intelligence Review},
  volume={58},
  number={6},
  pages={162},
  year={2025},
  publisher={Springer}
}

@article{CHENG2026152_survey_industrial_defect_detection,
title = {A comprehensive survey for real-world industrial surface defect detection: Challenges, approaches, and prospects},
journal = {Journal of Manufacturing Systems},
volume = {84},
pages = {152-172},
year = {2026},
issn = {0278-6125},
doi = {https://doi.org/10.1016/j.jmsy.2025.11.022},
url = {https://www.sciencedirect.com/science/article/pii/S0278612525002845},
author = {Yuqi Cheng and Yunkang Cao and Haiming Yao and Wei Luo and Cheng Jiang and Hui Zhang and Weiming Shen}
}

@article{REN2026103728_FII-DETR,
    author = {Ren, Kun and Li, Zhengzhen and Du, Yongping and Han, Honggui and Wu, Yufeng},
    title = {FII-DETR: Few-shot object detection with fully information interaction},
    year = {2026},
    issue_date = {Mar 2026},
    publisher = {Elsevier Science Publishers B. V.},
    address = {NLD},
    volume = {127},
    number = {PA},
    issn = {1566-2535},
    url = {https://doi.org/10.1016/j.inffus.2025.103728},
    doi = {10.1016/j.inffus.2025.103728},
    journal = {Inf. Fusion},
    month = jun,
    numpages = {14}
}

@article{zhan2024yolopx,
title = {YOLOPX: Anchor-free multi-task learning network for panoptic driving perception},
journal = {Pattern Recognition},
volume = {148},
pages = {110152},
year = {2024},
issn = {0031-3203},
doi = {https://doi.org/10.1016/j.patcog.2023.110152},
url = {https://www.sciencedirect.com/science/article/pii/S003132032300849X},
author = {Jiao Zhan and Yarong Luo and Chi Guo and Yejun Wu and Jiawei Meng and Jingnan Liu}
}

@article{shi2025progressive,
title = {Progressive class-aware instance enhancement for aircraft detection in remote sensing imagery},
journal = {Pattern Recognition},
volume = {164},
pages = {111503},
year = {2025},
issn = {0031-3203},
doi = {https://doi.org/10.1016/j.patcog.2025.111503},
url = {https://www.sciencedirect.com/science/article/pii/S0031320325001633},
author = {Tianjun Shi and Jinnan Gong and Jianming Hu and Yu Sun and Guangzhen Bao and Pengfei Zhang and others}
}

@ARTICLE{chen2025yoloms,
  author={Chen, Yuming and Yuan, Xinbin and Wang, Jiabao and Wu, Ruiqi and Li, Xiang and Hou, Qibin and others},
  journal={IEEE Transactions on Pattern Analysis and Machine Intelligence}, 
  title={YOLO-MS: Rethinking Multi-Scale Representation Learning for Real-Time Object Detection}, 
  year={2025},
  volume={47},
  number={6},
  pages={4240-4252},
  doi={10.1109/TPAMI.2025.3538473}}

@inproceedings{11092306_cvpr25,
  author={Bian, Jinghao and Feng, Mingtao and Dong, Weisheng and Wu, Fangfang and Luo, Jianqiao and Wang, Yaonan and Shi, Guangming},
  booktitle={2025 IEEE/CVF Conference on Computer Vision and Pattern Recognition (CVPR)}, 
  title={Feature Information Driven Position Gaussian Distribution Estimation for Tiny Object Detection}, 
  year={2025},
  volume={},
  number={},
  pages={30376-30386},
  doi={10.1109/CVPR52734.2025.02828}}

@article{fan2024deep_chestx,
  title={A deep-learning-based framework for identifying and localizing multiple abnormalities and assessing cardiomegaly in chest X-ray},
  author={Fan, Weijie and Yang, Yi and Qi, Jing and Zhang, Qichuan and Liao, Cuiwei and Wen, Li and others},
  journal={Nature Communications},
  volume={15},
  number={1},
  pages={1347},
  year={2024},
  publisher={Nature Publishing Group UK London}
}

@inproceedings{li2023large,
  author={Li, Yuxuan and Hou, Qibin and Zheng, Zhaohui and Cheng, Ming-Ming and Yang, Jian and Li, Xiang},
  booktitle={2023 IEEE/CVF International Conference on Computer Vision (ICCV)}, 
  title={Large Selective Kernel Network for Remote Sensing Object Detection}, 
  year={2023},
  volume={},
  number={},
  pages={16748-16759},
  doi={10.1109/ICCV51070.2023.01540}}

@inproceedings{meng2021conditional,
  author={Meng, Depu and Chen, Xiaokang and Fan, Zejia and Zeng, Gang and Li, Houqiang and Yuan, Yuhui and others},
  title={Conditional DETR for Fast Training Convergence}, 
  booktitle={2021 IEEE/CVF International Conference on Computer Vision (ICCV)}, 
  year={2021},
  volume={},
  number={},
  pages={3631--3640},
  doi={10.1109/ICCV48922.2021.00363}}

@inproceedings{yang2022querydet,
  author={Yang, Chenhongyi and Huang, Zehao and Wang, Naiyan},
  booktitle={2022 IEEE/CVF Conference on Computer Vision and Pattern Recognition (CVPR)}, 
  title={QueryDet: Cascaded Sparse Query for Accelerating High-Resolution Small Object Detection}, 
  year={2022},
  volume={},
  number={},
  pages={13658-13667},
  doi={10.1109/CVPR52688.2022.01330}}

@article{zhang2023idd,
    title = {IDD-Net: Industrial defect detection method based on Deep-Learning},
    journal = {Engineering Applications of Artificial Intelligence},
    volume = {123},
    pages = {106390},
    year = {2023},
    issn = {0952-1976},
    doi = {https://doi.org/10.1016/j.engappai.2023.106390},
    url = {https://www.sciencedirect.com/science/article/pii/S0952197623005742},
    author = {Zekai Zhang and Mingle Zhou and Honglin Wan and Min Li and Gang Li and Delong Han}
}

@article{sapkota2026object,
    title = {Object detection with multimodal large vision-language models: An in-depth review},
    journal = {Information Fusion},
    volume = {126},
    pages = {103575},
    year = {2026},
    issn = {1566-2535},
    doi = {https://doi.org/10.1016/j.inffus.2025.103575},
    url = {https://www.sciencedirect.com/science/article/pii/S1566253525006475},
    author = {Ranjan Sapkota and Manoj Karkee}
}

@article{lv2025lightweight,
title = {A lightweight hierarchical aggregation task alignment network for industrial surface defect detection},
journal = {Expert Systems with Applications},
volume = {263},
pages = {125727},
year = {2025},
issn = {0957-4174},
doi = {https://doi.org/10.1016/j.eswa.2024.125727},
url = {https://www.sciencedirect.com/science/article/pii/S0957417424025946},
author = {Shengping Lv and Tairan Liang and Kaibin Zhang and Shixin Jiang and Bin Ouyang and Quanzhou Li and Xiaoqing Li}
}

@ARTICLE{11297835,
  author={Cong, Runmin and Chen, Zhiyang and Fang, Hao and Kwong, Sam and Zhang, Wei},
  journal={IEEE Transactions on Pattern Analysis and Machine Intelligence}, 
  title={Breaking Barriers, Localizing Saliency: A Large-Scale Benchmark and Baseline for Condition-Constrained Salient Object Detection}, 
  year={2026},
  volume={48},
  number={4},
  pages={4167-4183},
  doi={10.1109/TPAMI.2025.3642893}}

@ARTICLE{11397189,
  author={Xia, Qiming and zheng, Longhui and Zhao, Shijia and Huang, Xun and Wu, Hai and Wen, Chenglu and Wang, Cheng},
  journal={IEEE Transactions on Pattern Analysis and Machine Intelligence}, 
  title={DOtA++: Unsupervisely and Collaboratively Detect Objects From Multi-Agent Observations With Multi-Modal Prior Constraints}, 
  year={2026},
  volume={48},
  number={7},
  pages={7467-7484},
  doi={10.1109/TPAMI.2026.3664863}}

@article{WEI2026103906,
title = {SAM-Swin: SAM-driven dual-swin transformers with adaptive lesion enhancement for Laryngo-Pharyngeal tumor detection},
journal = {Medical Image Analysis},
volume = {109},
pages = {103906},
year = {2026},
issn = {1361-8415},
doi = {https://doi.org/10.1016/j.media.2025.103906},
url = {https://www.sciencedirect.com/science/article/pii/S1361841525004529},
author = {Jia Wei and Yun Li and Xiaomao Fan and Wenjun Ma and Meiyu Qiu and Hongyu Chen and Wenbin Lei}
}

@article{ZHANG2022102415,
title = {DDTNet: A dense dual-task network for tumor-infiltrating lymphocyte detection and segmentation in histopathological images of breast cancer},
journal = {Medical Image Analysis},
volume = {78},
pages = {102415},
year = {2022},
issn = {1361-8415},
doi = {https://doi.org/10.1016/j.media.2022.102415},
url = {https://www.sciencedirect.com/science/article/pii/S1361841522000676},
author = {Xiaoxuan Zhang and Xiongfeng Zhu and Kai Tang and Yinghua Zhao and Zixiao Lu and Qianjin Feng}
}

@article{zhang2022focaleiou,
title = {Focal and efficient IOU loss for accurate bounding box regression},
journal = {Neurocomputing},
volume = {506},
pages = {146-157},
year = {2022},
issn = {0925-2312},
doi = {https://doi.org/10.1016/j.neucom.2022.07.042},
url = {https://www.sciencedirect.com/science/article/pii/S0925231222009018},
author = {Yi-Fan Zhang and Weiqiang Ren and Zhang Zhang and Zhen Jia and Liang Wang and Tieniu Tan}
}

@inproceedings{alpha_iou,
author = {He, Jiabo and Erfani, Sarah and Ma, Xingjun and Bailey, James and Chi, Ying and Hua, Xian-Sheng},
title = {Alpha-IoU: a family of power intersection over union losses for bounding box regression},
year = {2021},
isbn = {9781713845393},
publisher = {Curran Associates Inc.},
address = {Red Hook, NY, USA},
booktitle = {Proceedings of the 35th International Conference on Neural Information Processing Systems},
articleno = {1547},
numpages = {13},
series = {NIPS '21}
}

@article{tong2023wiseiou,
  title={Wise-IoU: bounding box regression loss with dynamic focusing mechanism},
  author={Tong, Zanjia and Chen, Yuhang and Xu, Zewei and Yu, Rong},
  journal={arXiv preprint arXiv:2301.10051},
  year={2023}
}

@inproceedings{yang2025pinwheel,
author = {Yang, Jiangnan and Liu, Shuangli and Wu, Jingjun and Su, Xinyu and Hai, Nan and Huang, Xueli},
title = {Pinwheel-shaped convolution and scale-based dynamic loss for infrared small target detection},
year = {2025},
isbn = {978-1-57735-897-8},
publisher = {AAAI Press},
url = {https://doi.org/10.1609/aaai.v39i9.32996},
doi = {10.1609/aaai.v39i9.32996},
booktitle = {Proceedings of the Thirty-Ninth AAAI Conference on Artificial Intelligence and Thirty-Seventh Conference on Innovative Applications of Artificial Intelligence and Fifteenth Symposium on Educational Advances in Artificial Intelligence},
articleno = {1023},
numpages = {9},
series = {AAAI'25/IAAI'25/EAAI'25}
}

@inproceedings{wang2025yoloe,
  author={Wang, Ao and Liu, Lihao and Chen, Hui and Lin, Zijia and Han, Jungong and Ding, Guiguang},
  booktitle={2025 IEEE/CVF International Conference on Computer Vision (ICCV)}, 
  title={YOLOE: Real-Time Seeing Anything}, 
  year={2025},
  volume={},
  number={},
  pages={24591-24602},
  doi={10.1109/ICCV51701.2025.02280}}

@inproceedings{li2023lite_detr,
  author={Li, Feng and Zeng, Ailing and Liu, Shilong and Zhang, Hao and Li, Hongyang and Zhang, Lei and others},
  booktitle={2023 IEEE/CVF Conference on Computer Vision and Pattern Recognition (CVPR)}, 
  title={Lite DETR : An Interleaved Multi-Scale Encoder for Efficient DETR}, 
  year={2023},
  volume={},
  number={},
  pages={18558-18567},
  doi={10.1109/CVPR52729.2023.01780}}

@inproceedings{zhao2024msdetr,
  author={Zhao, Chuyang and Sun, Yifan and Wang, Wenhao and Chen, Qiang and Ding, Errui and Yang, Yi and others},
  booktitle={2024 IEEE/CVF Conference on Computer Vision and Pattern Recognition (CVPR)}, 
  title={MS-DETR: Efficient DETR Training with Mixed Supervision}, 
  year={2024},
  volume={},
  number={},
  pages={17027-17036},
  doi={10.1109/CVPR52733.2024.01611}}

@inproceedings{set_2025_CVPR,
  author={Sun, Huixin and Wang, Runqi and Li, Yanjing and Yang, Linlin and Lin, Shaohui and Cao, Xianbin and Zhang, Baochang},
  booktitle={2025 IEEE/CVF Conference on Computer Vision and Pattern Recognition (CVPR)}, 
  title={SET: Spectral Enhancement for Tiny Object Detection}, 
  year={2025},
  volume={},
  number={},
  pages={4713-4723},
  doi={10.1109/CVPR52734.2025.00444}}

@article{zhang2022dino,
  title={Dino: Detr with improved denoising anchor boxes for end-to-end object detection},
  author={Zhang, Hao and Li, Feng and Liu, Shilong and Zhang, Lei and Su, Hang and Zhu, Jun and Ni, Lionel M and Shum, Heung-Yeung},
  journal={arXiv preprint arXiv:2203.03605},
  year={2022}
}
\end{document}